\documentclass[10pt,journal,compsoc]{IEEEtran}

\usepackage{cite}
\ifCLASSINFOpdf
   \usepackage[pdftex]{graphicx}
\else
\fi
\usepackage{amsmath}
\usepackage{algorithmic}
\usepackage{algorithm}

\usepackage{array}
\newcolumntype{x}[1]{>{\centering\arraybackslash\hspace{0pt}}p{#1}}

\ifCLASSOPTIONcompsoc
 \usepackage[caption=false,font=footnotesize,labelfont=sf,textfont=sf]{subfig}
\else
 \usepackage[caption=false,font=footnotesize]{subfig}
\fi

\usepackage{multirow}
\usepackage{arydshln}
\usepackage{amssymb}
\usepackage{enumitem}

\begin{document}
%
\title{Predicting Biomedical Interactions with Higher-Order Graph Convolutional Networks}
%
%
%
%

\author{
Kishan~KC,
        Rui Li,
        Feng Cui,
        and~Anne R. Haake
\IEEEcompsocitemizethanks{\IEEEcompsocthanksitem K. KC, R. Li, and A. Haake are with the Department
of Computing and Information Sciences, Rochester Institute of Technology, Rochester,
NY, 14623.\protect\\
E-mail: \{kk3671, arhics, rxlics\}@rit.edu
\IEEEcompsocthanksitem F. Cui is with Thomas H. Gosnell School of Life Sciences, Rochester Institute of Technology, Rochester,
NY, 14623. E-mail: fxcsbi@rit.edu
}
}

%
%

\markboth{Journal of \LaTeX\ Class Files,~Vol.~14, No.~8, August~2015}%
{Shell \MakeLowercase{\textit{et al.}}: Bare Advanced Demo of IEEEtran.cls for IEEE Computer Society Journals}
%



\IEEEtitleabstractindextext{%
\begin{abstract}
Biomedical interaction networks have incredible potential to be useful in the prediction of biologically meaningful interactions, identification of network biomarkers of disease, and the discovery of putative drug targets. Recently, graph neural networks have been proposed to effectively learn representations for biomedical entities and achieved state-of-the-art results in biomedical interaction prediction. These methods only consider information from immediate neighbors but cannot learn a general mixing of features from neighbors at various distances. In this paper, we present a higher-order graph convolutional network (HOGCN) to aggregate information from the higher-order neighborhood for biomedical interaction prediction. Specifically, HOGCN collects feature representations of neighbors at various distances and learns their linear mixing to obtain informative representations of biomedical entities. Experiments on four interaction networks, including protein-protein, drug-drug, drug-target, and gene-disease interactions, show that HOGCN achieves more accurate and calibrated predictions. HOGCN performs well on noisy, sparse interaction networks when feature representations of neighbors at various distances are considered. Moreover, a set of novel interaction predictions are validated by literature-based case studies.
\end{abstract}

}

\maketitle

\IEEEdisplaynontitleabstractindextext

%
\IEEEpeerreviewmaketitle

\ifCLASSOPTIONcompsoc
\IEEEraisesectionheading{\section{Introduction}\label{sec:introduction}}
\else
\section{Introduction}
\label{sec:introduction}
\fi

%
%
%
%
\IEEEPARstart{A} biological system is a complex network of various molecular entities such as genes, proteins, and other biological molecules linked together by the interactions between these entities. The complex interplay between various molecular entities can be represented as interaction networks with molecular entities as nodes and their interactions as edges. Such a representation of a biological system as a network provides a conceptual and intuitive framework to investigate and understand direct or indirect interactions between different molecular entities in a biological system. Study of such networks lead to system-level understanding of biology~\cite{cowen2017network} and discovery of novel interactions including protein-protein interactions (PPIs)~\cite{luck2020reference}, drug-drug interactions (DDIs)~\cite{zitnik2018modeling}, drug-target interactions (DTIs)~\cite{luo2017network} and gene-disease associations (GDIs)~\cite{agrawal2018large}. 

Recently, the generalization of deep learning to the network-structured data~\cite{bronstein2017geometric} has shown great promise across various domains such as social networks~\cite{gnn@social}, recommendation systems~\cite{gnn@recommendation}, chemistry~\cite{gilmer2017neural}, citation networks~\cite{kipf2016variational}. These approaches are under the umbrella of graph convolutional networks (GCNs). GCNs repeatedly aggregate feature representations of immediate neighbors to learn the informative representation of the nodes for link prediction. Although GCN based methods show great success in biomedical interaction prediction~\cite{Yue2020, zitnik2018modeling}, the issue with such methods is that they only consider information from immediate neighbors. SkipGNN~\cite{huang2020skipgnn} leverages skip graph to aggregate feature representations from direct and second-order neighbors and demonstrated improvements over GCN methods in biomedical interaction prediction. However, SkipGNN cannot be applied to aggregate information from higher-order neighbors and thus fail to capture information that resides farther away from a particular interaction~\cite{abu2019mixhop}. 

To address the challenge, we propose an end-to-end deep graph representation learning framework named higher-order graph convolutional networks (HOGCN) for predicting interactions between pairs of biomedical entities. HOGCN learns a representation for every biomedical entity using an interaction network structure $\mathcal{G}$ and/or features $X$. In particular, we define a higher-order graph convolution (HOGC) layer where the feature representations from $k$-order neighbors are considered to obtain the representation of biomedical entities. The layer can thus learn to mix feature representations of neighbors at various distances for interaction prediction. Furthermore, we define a bilinear decoder to reconstruct the edges in the input interaction network $\mathcal{G}$ by relying on feature representations produced by HOGC layers. The encoder-decoder approach makes HOGCN an end-to-end trainable model for interaction prediction.

We compare HOGCN's performance with that of state-of-the-art network similarity-based methods~\cite{kovacs2019network}, network embedding methods~\cite{perozzi2014deepwalk, grover2016node2vec}, and graph convolution-based methods~\cite{kipf2016semi, kipf2016variational, huang2020skipgnn} for biomedical link prediction. We experiment with various interaction datasets and show that our method makes accurate and calibrated predictions. HOGCN outperforms alternative methods based on network embedding by up to 30\%. Furthermore, HOGCN outperforms graph convolution-based methods by up to 6\%, alluding to the benefits of aggregating information from higher-order neighbors. 

We perform a case study on the DDI network and observe that aggregating information from higher-order neighborhood allows HOGCN to learn meaningful representation for drugs. Moreover, literature-based case studies illustrate that the novel predictions are supported by evidence, suggesting that HOGCN can identify interactions that are highly likely to be a true positive. 

In summary, our study demonstrates the ability of HOGCN to identify potential interactions between biomedical entities and opens up the opportunities to use the biological and physicochemical properties of biomedical entities for a follow-up analysis of these interactions.

\section{Related works}
With the increasing coverage of the interactome, various network-based approaches have been proposed to exploit already available interactions to predict missing interactions~\cite{ahmad2020missing, keskin2016predicting, kishan2019gne,kovacs2019network}. These methods can be broadly classified into (1) network similarity-based methods (2) network embedding methods (3) graph convolution-based methods. We next summarize these categories of methods for biomedical interaction prediction. 

Given a network of known interactions, various similarity metrics are used to measure the similarity between the biomedical entities~\cite{ahmad2020missing} with an assumption that higher similarity indicates interaction. Triadic closure principle (TCP) has been explored in biomedical interaction prediction with the hypothesis that biomedical entities with common interaction partners are likely to interact with each other~\cite{keskin2016predicting}. TCP relies on a common neighbor algorithm to count the number of shared neighbors between the nodes and is quantified by $A^2$ where $A$ is the adjacency matrix. Recently, L3 heuristic~\cite{kovacs2019network} shows the common neighbor hypothesis fails for most protein pairs in PPI prediction and proposes to consider nodes that are similar to the neighbors of the nodes and can be quantified by $A^3$. This indicates that higher-order neighbors are important for interaction prediction.

Next, network embedding methods embed the existing networks to low-dimensional space that preserves the structural proximity such that the nodes in the original network can be represented as low-dimensional vectors. Deepwalk~\cite{perozzi2014deepwalk} is a popular approach that generates the truncated random walks in the network and defines a neighborhood for each node as a set of nodes within a window of size $k$ in each random walk. Similarly, node2vec performs a biased random walk by balancing the breadth-first and depth-first search in the network. The random walks generated by these methods can be considered as a combination of nodes from various order of neighborhoods such as 1-hop to $k$-hop neighborhood. In other words, DeepWalk and node2vec learn the embeddings for the nodes in the network from the combination of $A^1, A^2, A^3, \ldots A^k$ where $A^i$ is the $i^{th}$ power of the adjacency matrix. These embeddings can then be fed into a classifier to predict the interaction probability. These methods are only limited to the structure of the biomedical networks and cannot incorporate additional information about the biomedical entities. Also, they cannot learn the feature difference between nodes at various distances.

Furthermore, graph convolution-based methods use a message-passing mechanism to receive and aggregate information from neighbors to generate representations for the nodes in the network. Graph convolutional networks (GCNs)~\cite{kipf2016semi} and variational graph convolutional autoencoder (VGAE)~\cite{kipf2016variational} aggregate feature representation from immediate neighbors to learn the representation of biomedical entities in an end-to-end manner using link prediction objective. These methods are only limited to the average pooling of the neighborhood features~\cite{abu2019mixhop}. SkipGNN~\cite{huang2020skipgnn} therefore proposes to use skip similarity between the biomedical entities to aggregate information from second-order neighbors. However, these methods cannot aggregate feature representations from higher-order neighbors and also cannot learn feature differences between neighbors at various distances.

\section{Preliminaries}
A biomedical network is defined as $\mathcal{G} = (\mathcal{V},\mathcal{E},X)$ where $\mathcal{V}$ denotes the set of nodes representing biomedical entities (e.g. proteins, genes, drugs, diseases) and $|\mathcal{V}|$ denotes the number of nodes. $\mathcal{E} \subseteq (\mathcal{V} \times \mathcal{V})$ denotes a set of interactions between biomedical entities. $X \in \mathbb{R}^{|\mathcal{V}| \times F}$ is the features of biomedical entities where $F$ is the dimension of features. 

Let $A$ denote the adjacency matrix of $\mathcal{G}$, where $A_{ij}$ indicates an edge between nodes $v_i$ and $v_j$. 
We assume the case of binary adjacency matrix $A_{ij}\in \{0, 1\}^{n \times n}$ where $A_{ij}$ represents the existence of edge between the nodes $v_i$ and $v_j$, indicating the presence of the experimental evidence for their interaction (i.e. $A_{ij} = 1$) or the absence of the experimental evidence for their interaction (i.e. $A_{ij} = 0$). Note that the same notation of adjacency matrix can be used to represent weighted graphs such that $A_{ij} = [0, 1]$. Table~\ref{tab_notations} shows the notations and their definitions used in the paper.
\begin{table}[htb]
\caption{Terms and notations}
\label{tab_notations}
\begin{tabular}{ll}
\hline Notation & Definition \\
\hline
$\mathcal{G}:\{\mathcal{V}, \mathcal{E}, X\}$ & Graph with nodes $\mathcal{V}$, edges $\mathcal{E}$ and features $X$\\
$\mathcal{E}^{\prime}$ & Test edges \\
$A \in \mathbb{R}^{|\mathcal{V}| \times |\mathcal{V}|}$ & Adjacency matrix of graph $\mathcal{G}$\\
$D \in \mathbb{R}^{|\mathcal{V}| \times |\mathcal{V}|}$ & Degree matrix with $D_{ii} = \sum_i{A_{ij}}$\\
$I \in \mathbb{R}^{|\mathcal{V}| \times |\mathcal{V}|}$ & Identity matrix \\
$\hat{A} \in \mathbb{R}^{|\mathcal{V}| \times |\mathcal{V}|}$ & Symmetrically normalized adjacency matrix\\
$X \in \mathbb{R}^{|\mathcal{V}| \times F}$ & $F$-dimensional feature matrix \\
$A_{i j} \in\{0,1\}$ & Ground-truth interaction between nodes $i$ and $j$ \\
$p_{i j} \in[0,1]$ & Probability of interaction between nodes $i$ and $j$ \\
$ Z\in \mathbb{R}^{|\mathcal{V}| \times d^*}$ & Final node embeddings \\
$W_{(l)}^{(i)}$ & Weight matrix for $i^{\text{th}}$ adjacency power for layer $l$ \\
$L$ & Number of HOGC layers \\
$T$ & Number of training epochs \\
$k$ & The order of neighborhood \\
$P$ & A set of integer adjacency powers $P =\{0, 1, \ldots, k \}$\\
$O_j^{(l)}$ & Representation of neighbors at distance $j$ in layer $l$\\
\hline
\end{tabular}
\end{table}

\textbf{Problem Statement.} (Biomedical interaction prediction) Given a biomedical interaction network $\mathcal{G} = (\mathcal{V}, \mathcal{E}, X)$ and the set of potential biomedical interactions $\mathcal{E}^{\prime}$, we aim to learn a interaction prediction model $f$ to predict the interaction probabilities of $\mathcal{E}^{\prime}$, $f: \mathcal{E}^{\prime} \rightarrow [0, 1]$.

\begin{figure*}[!t]
\centering
\includegraphics[width=\textwidth]{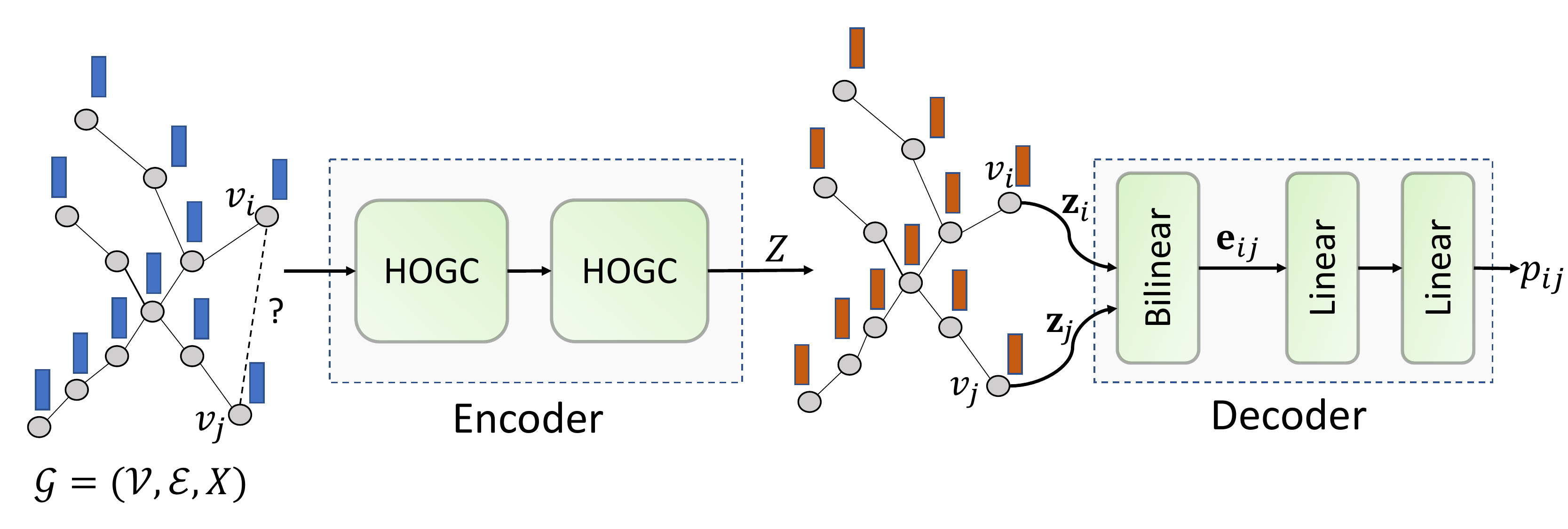}
\caption{Block diagram of proposed HOGCN model with 2 HOGC layers. Given a biomedical interaction network $\mathcal{G}$ with initial features $X$ for biomedical entities, the encoder mixes the feature representation of neighbors at various distances and learn final representation $Z$. The decoder takes the representation $\mathbf{z}_i$ and $\mathbf{z}_j$ of nodes $v_i$ and $v_j$ to learn the representation $\mathbf{e}_{ij}$ for the edge (denoted by $?$) and predict probability $p_{ij}$ of its existence.}
\label{block_diagram}
\end{figure*}

\subsection{Message Passing}\label{message_passing}
Given a biomedical network $\mathcal{G}$, message passing algorithms learn the representation of biomedical entities in the network by aggregating information from immediate neighbors~\cite{gilmer2017neural}. Additional information about biomedical entities can be used to initialize the feature matrix $X$. These algorithms involve the message passing step in which each biomedical entity sends its current representation to, and aggregates incoming messages from its immediate neighbors. Representation for each biomedical entity can be obtained after $L$ steps of message passing and feature aggregation. However, such message passing operation is limited to average pooling of features from immediate neighbors and thus is unable to learn feature differences among neighbors at various distances~\cite{abu2019mixhop}. 

Neighborhood nodes at various distances provide network structure information at different granularities~\cite{benson2018simplicial, cao2015grarep, zhu2018high, HONRL, rossi2018hone}. Taking $k$-hop neighborhoods into consideration, we aim at aggregating information from various distances at every message passing step.  Different powers of adjacency matrices such as $A^1, A^2, A^3, \ldots, A^k$ provide information about the network structure at different scales. Higher-order message passing operations can therefore learn to mix their representations using various powers of the adjacency matrix at each message passing step.

\subsection{Graph Convolutional Networks (GCNs)}
Graph convolutional networks (GCNs) are the generalization of convolution operation from regular grids such as images or texts to graph structured data~\cite{bronstein2017geometric, wu2020comprehensive}. The key idea of GCNs is to learn the function to generate the node's representation by repeatedly aggregating information from immediate neighbors. The graph convolutional layer is defined as:
\begin{equation}
    H^{(l)} = \sigma (\hat{A}H^{(l-1)}W^{(l)})
\end{equation}
where $H^{(l-1)}$ and $H^{(l)}$ are the input and output activations, $W^{(l)}$ is a trainable weight matrix of the layer $l$, $\sigma$ is the element-wise activation, and $\hat{A}$ is a symmetrically normalized adjacency matrix with self-connections $\hat{A} = D^{-\frac{1}{2}}(A + I_{|\mathcal{V}|})D^{-\frac{1}{2}}$. A GCN model with $L$ layers is then defined as:
\begin{equation*}
 H^{(l)} = \begin{cases}
X & \text{if $l = 0$}\\
\sigma (\hat{A}H^{(l-1)}W^{(l)}) &\text{if $l \in [1, \ldots, L]$}
\end{cases}
\end{equation*}
and $H^{(L)}$ can be used to predict the probability of interactions between biomedical entities.

\section{Higher-order Graph Convolution Network (HOGCN)}
In this work, we develop a higher-order graph convolutional network (HOGCN) that takes an interaction network $\mathcal{G}$ as input and reconstruct the edges in the interaction network (Fig.~\ref{block_diagram}). HOGCN has two main components:
\begin{itemize}
    \item \textbf{Encoder}: a higher-order graph convolution encoder that operates on an interaction graph $\mathcal{G}$ and produces representations for biomedical entities by aggregating features from the neighborhood at various distances and
    \item \textbf{Decoder}: a bilinear decoder that relies on these representations to reconstruct the interactions in $\mathcal{G}$.
\end{itemize}

\subsection{Higher-Order Graph Encoder}
We first describe the higher-order graph encoder, that operates on an undirected interaction graph $\mathcal{G}= (\mathcal{V}, \mathcal{E}, X)$ and learns the representations for biomedical entities.

We develop an encoder with higher-order Graph Convolution (HOGC) layer to mix feature representations from neighbors at $k$-distances. Specifically, HOGC layer considers the neighborhood information at different granularities and is defined as:
\begin{equation}\label{mixhop}
H^{(l)}=\underset{j \in P}{\|} \sigma\left(\widehat{A}^{j} H^{(l-1)} W_{j}^{(l)}\right)
\end{equation}
where $P$ is a set of integer adjacency powers, $\widehat{A}^{j}$ denotes the adjacency matrix $\widehat{A}$ multiplied $j$ times, and $\|$ denotes column-wise concatenation~\cite{abu2019mixhop}. Graph convolutional network~\cite{kipf2016semi} only considers the $1^{\text{st}}$ power of adjacency matrix and can be exactly recovered by setting $P=\{1\}$ in Equation~(\ref{mixhop}). Similarly, SkipGNN~\cite{huang2020skipgnn} considers direct and skip similarity and can be recovered by setting $P = \{1, 2\}$ in Equation~(\ref{mixhop}).

\begin{figure}[htb]
\centering
\includegraphics[width=\linewidth]{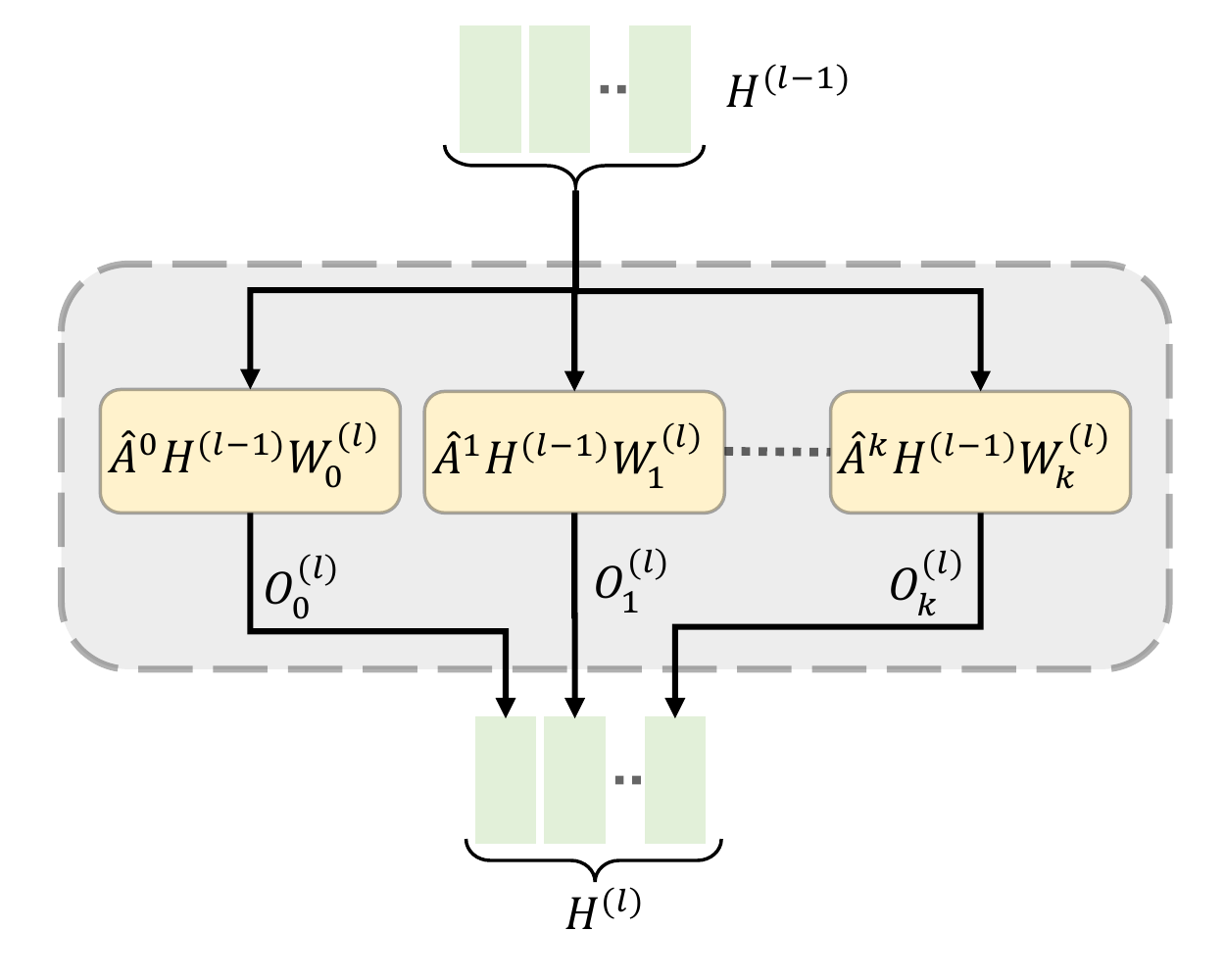}
\caption{High-order graph convolution (HOGC) Layer with $P =\{0,\ldots, k\}$. The feature representation $H^{(l)}$ is a linear combination of the neighbors $\widehat{A}^{j} H^{(l-1)}$ at multiple distances $j$. $O_{j}^{(l)}$ represents feature representation of neighbors at $j$ distances for layer $l$.}
\label{fig_gcn_layer}
\end{figure}

Fig.~\ref{fig_gcn_layer} shows a HOGC layer with $P = \{0, 1,\ldots, k\}$ where $k$ is maximum order of neighborhood considered in each HOGC layer. If $k=0$, HOGC layer only considers the features of the biomedical entities and can capture the feature similarity between various biological entities. This is equivalent to a fully connected network with features of biomedical entities as input. For the HOGC layer, $\widehat{A}^0$ is the identity matrix $I_{|\mathcal{V}|}$ where $|\mathcal{V}|$ is the number of nodes in the network. This allows the HOGC layer to learn the transformation of node features separately and mix it with feature representations from neighbors. 

The maximum order of neighborhood $k$ and the number of trainable weight matrices $|P|$, one per each adjacency power, can vary across layers. However, we set the same $k$ for neighborhood aggregation and the same dimension $d$ for all the weight matrices across all layers.
 
Neighborhood features from different adjacency powers $j \in \{0, 1, \ldots, k\}$ at layer $(l-1)$ are column-wise concatenated to obtain feature representation $H^{(l-1)}$. As shown in Fig~\ref{fig_gcn_layer}, weight $W_{(\cdot)}^{(l)}$ at layer $l$ can learn the arbitrary linear combination of the concatenated features to obtain $H^{(l)}$. Specifically, the layer can assign different coefficients to different columns in the concatenated matrix. For instance, the layer can assign positive coefficients to the columns produced by certain power of $\widehat{A}$ and assign negative coefficients to other powers. This allows the model to learn feature differences among neighbors at various distances. We apply $L$ HOGC layers to learn the latent representation $Z \in \mathbb{R}^{|\mathcal{V}| \times d^*}$ for biomedical entities in the network, where $d^* = d \times |P|$  and $d$ is the dimension of node's representation for each adjacency power.

\subsection{Interaction Decoder}
We introduced the encoder based on HOGC layers that learns feature representation $Z$ for biomedical entities by mixing neighborhood information at multiple distances. Next, we discuss the decoder that reconstructs the interactions in $\mathcal{G}$ based on the representation $Z$.

We adopt a bilinear layer to fuse the representation of biomedical entities $v_i$ and $v_j$ and learn the edge representation $\mathbf{e}_{ij}$. More precisely, we define a simple bilinear layer that takes the representation $\mathbf{z}_i$ and $\mathbf{z}_j$ as input:
\begin{equation}
    \mathbf{e}_{ij} = \mathrm{ELU}(\mathbf{z}_i^T W_b \mathbf{z}_j + b)
\end{equation}
where $W_b \in \mathbb{R}^{d \times d^* \times d^*}$ represents the learnable fusion matrix, $\mathbf{e}_{ij}$ is the representation of edge $e_{ij}$ between nodes $v_i$ and $v_j$, $b$ denotes the bias of the bilinear layer. $\mathrm{ELU}$ is non-linearity.

The edge representation $\mathbf{e}_{ij}$ is then fed into 2-layered fully connected neural network to predict probability ${p}_{ij}$ for edge $e_{ij}$:
\begin{equation}
    p_{ij} = \text{sigmoid}(\mathrm{FC}_2(\mathrm{ELU}(\mathrm{FC}_1(\mathbf{e}_{ij}))))
\end{equation}
where $\mathrm{FC}_1(\mathbf{e}_{ij}) = W_1\cdot \mathbf{e}_{ij} + b_1$ denotes fully connected layer with weight $W_1$ and bias $b_1$, $p_{ij}$ represents the probability that biomedical entities $v_i$ and $v_j$ interact.

So far, we have discussed the encoder and decoder of our proposed approach.
Next, we describe the training procedure of our proposed HOGCN model. In particular, we explain how to optimize the trainable neural network weights in an end-to-end manner.

\subsection{Training HOGCN}
During HOGCN training, we employ binary cross entropy loss to optimize the model parameters
\begin{equation}\label{loss}
\mathcal{L}(v_i, v_j)=- A_{i j} \log (p_{i j}) -\left(1-A_{i j}\right) \log \left(1-p_{i j}\right)
\end{equation} 
and encourage the model to assign higher probability to observed interactions $(v_i, v_j)$ than to randomly selected non-interactions. $p_{ij}$ is the predicted interaction probability between $v_i$ and $v_j$ and $A_{ij}$ denotes the ground-truth interaction label between these nodes. The final loss function considering all interactions is 
\begin{equation}\label{total_loss}
\mathcal{L}=\sum_{(i, j) \in \mathcal{E}} \mathcal{L}(v_i, v_j)
\end{equation}
We follow an end-to-end approach to jointly optimize over all trainable parameters and backpropagate the gradients through encoder and decoder of HOGCN.

\subsubsection{Algorithm}
HOGCN leverages biomedical network structure $A$ along with additional information about biomedical entities as the initial feature representation $X$. In this paper, we initialize the initial features $X$ to be one-hot encoding \textit{i.e.} $I_{|\mathcal{V}|}$. The feature matrix $X$ can be initialized with properties of biomedical entities or pre-trained embeddings from other network-based approaches such as DeepWalk, node2vec.

\begin{algorithm}[!htb]
\caption{Training of HOGCN for biomedical interaction prediction}
\begin{algorithmic}[1]
\label{alg_training}
\STATE \textbf{Inputs:} $\hat{A}$, $X$, $k$
\STATE $H^{(0)} = X$
\FOR{$t=1$ to $T$}
\STATE Sample mini-batch of training edges and their interaction labels
\FOR{$l=1$ to $L$}
\STATE $B := H^{(l-1)}$
\FOR{$j=1$ to $k$}
\STATE $B := \hat{A}B$
\STATE $O_j^{(l)} := BW^{(l)}_j$
\ENDFOR
\STATE $H^{(l)} := \underset{j \in P}{\|} O_j^{(l)}$
\ENDFOR
\STATE $Z := H^{(L)}$
\STATE $p_{ij} := \textbf{Interaction Decoder}(Z)$
\STATE Compute loss in (\ref{total_loss})
\STATE Update model parameters via gradient descent
\ENDFOR
\end{algorithmic}
\end{algorithm}

Given an adjacency matrix $A$ and the initial node representations $X$, higher-order neighborhood indicated by the higher power of the adjacency matrix is iteratively computed that makes the model more efficient. By adopting right-to-left multiplication, for instance, we can calculate $\hat{A}^3H^{(i)}$ as $\hat{A}(\hat{A}(\hat{A}H^{(i)}))$ (Line $8$ in Algorithm~\ref{alg_training}). Representation $O_j^{(l)}$ learned for the neighborhood at $j$ distances are concatenated to obtain the representation $H^{(l)}$ as shown in Fig.~\ref{fig_gcn_layer} (Line 11 in Algorithm~\ref{alg_training}). After passing through $L$ HOGC layers, we obtain the final representation $Z$ for biomedical entities. With the final representations $Z$ and the mini-batch of training edges, we retrieve the embeddings for the nodes in training edges and feed them into the interaction decoder to compute their interaction probabilities.

The parameters of HOGCN are optimized with a binary cross-entropy loss (Equation~(\ref{total_loss})) in an end-to-end manner. Given two biomedical entities $v_i$ and $v_j$, the trained model can predict the probability of their interactions.


\section{Experimental design}
We view the problem of biomedical interaction prediction as solving a link prediction task on an interaction network. We consider various interaction datasets and compare our proposed method with the state-of-the-art methods. 
\subsection{Datasets}
We conduct interaction prediction experiments on four publicly-available biomedical network datasets: 
\begin{itemize}
    \item \textbf{BioSNAP-DTI}~\cite{zitnik2018biosnap}: DTI network contains 15,139 drug-target interactions between 5,018 drugs and 2,325 proteins.
    \item  \textbf{BioSNAP-DDI}~\cite{zitnik2018biosnap}: DDI network contains 48,514 drug-drug interactions between 1,514 drugs extracted from drug labels and biomedical literature.
    \item  \textbf{HuRI-PPI}~\cite{luck2020reference}: HI-III human PPI network contains 5,604 proteins and 23,322 interactions generated by multple orthogonal high-throughput yeast two-hybrid screens. 
    \item \textbf{DisGeNET-GDI}~\cite{pinero2020disgenet}: GDI network consists of 81,746 interactions between 9,413 genes and 10,370 diseases curated from GWAS studies, animal models and scientific literature.
\end{itemize}
Table~\ref{tab:datasets} provides summary of datasets used in our experiments. 
We provide the number of interactions used for training, validation, and testing for each interaction datasets. Also, the table includes the average number of interactions for each biomedical entity which can be computed as $\frac{2|\mathcal{E}|}{|\mathcal{V}|}$.

\begin{table*}[htb]
    \centering
    \renewcommand{\arraystretch}{1.2}
    \caption{Summary of the datasets used in our experiments.}
    \label{tab:datasets}
    \begin{tabular}{ccccccc}
    \hline
    \multirow{2}{*}{Dataset} & \multirow{2}{*}{\# Nodes} &\multicolumn{4}{c}{\# Edges} &  \multirow{2}{*}{Avg. node degree} \\ 
    \cline{3-6}
     &  & Training (70\%) & Validation (10\%) & Testing (20\%) & Total & \\
    \hline
    DTI & 5,018 drugs, 2,325 proteins & 10,597 & 1,514 &  3,028 & 15,139 & 4.12\\
    DDI & 1,514 drugs & 33,960 & 4,852 & 9,702 & 48,514 & 64.09  \\
    PPI &  5,604 proteins  & 16,326 & 2,332 & 4,664 &  23,322 & 8.32 \\
    GDI & 9,413 genes, 10,370 diseases &  57,222 & 8,175 & 16,349 & 81,746 & 8.26\\
    \hline
    \end{tabular}
\end{table*}

\subsection{Baselines}
We compare our proposed model with the following network-based baselines for interaction prediction:
\begin{itemize}
    \item {network similarity-based methods}
    \begin{itemize}[leftmargin=*]
        \item \textbf{L3}~\cite{kovacs2019network} counts the number of paths with length-3 normalized by the degree for all the node pairs.  
    \end{itemize}
    \item {Network embedding methods}
    \begin{itemize}[leftmargin=*]
        \item \textbf{DeepWalk}~\cite{perozzi2014deepwalk} performs truncated random walk exploring the network neighborhood of nodes and applies skip-gram model to learn the $d$-dimensional embedding for each node in the network. Node features are concatenated to form edge representation and train a logistic regression classifier.  
        \item \textbf{node2vec}~\cite{grover2016node2vec} extends DeepWalk by running biased random walk based on breadth/depth-first search to capture both local and global network structure. 
    \end{itemize}
    \item {Graph convolution-based methods}
    \begin{itemize}[leftmargin=*]
        \item  \textbf{VGAE}~\cite{kipf2016variational} uses graph convolutional encoder with two GCN layers to learn representation for each node in the network and adopts inner product decoder to reconstruct adjacency matrix.
        \item \textbf{GCN}~\cite{kipf2016semi} uses normalized adjacency matrix to learn node representations. The representation for nodes are concatenated to form feature representation for the edges and fully connected layer use these edge representation to reconstruct edges, similar to HOGCN. Setting $P = \{1\}$ in our proposed HOGCN is equivalent to GCN.
        \item  \textbf{SkipGNN}~\cite{huang2020skipgnn} learns the node embeddings by combining direct and skip similarity between nodes. Setting $P = \{1, 2\}$ in our proposed HOGCN is equivalent to SkipGNN.
    \end{itemize}
\end{itemize}

\subsection{Experimental setup}
We split the interaction dataset into training, validation, and testing interactions in a ratio of 7:1:2 as shown in Table~\ref{tab:datasets}. Since the available interactions are positive samples, the negative samples are generated by randomly sampling from the complement set of positive examples. Five independent runs of the experiments with different random splits of the dataset are conducted to report the prediction performance. We use (1) area under the precision-recall curve (AUPRC) and (2) area under the receiver operating characteristics (AUROC) as the evaluation metrics. With these evaluation metrics, we expect positive interactions to have higher interaction probability compared to negative interactions. So, the higher value of AUPRC and AUROC indicates better performance. 

We implement HOGCN using PyTorch~\cite{paszke2019pytorch} and perform all experiments on a single NVIDIA GeForce RTX 2080Ti GPU. We construct a $2$-layered HOGC network with $k=3$ for each layer. At each HOGC layer, the node mixes the feature representations from neighbors at distances $P = \{0, 1, 2 $ and $3\}$. The dimension of all weight matrices in HOGC layers is set to $d=32$. All the weight matrices are initialized using Xavier initialization~\cite{glorot2010understanding}. We train our model using mini-batch gradient descent with Adam optimizer~\cite{kingma2015adam} for a maximum of 50 epochs, with a fixed learning rate of $5 \times 10^{-4}$. We set the mini-batch size to 256 and the dropout probability~\cite{srivastava2014dropout} to 0.1 for all layers. Early stopping is adopted to stop training if validation performance does not improve for 10 epochs. The dimension of the edge feature $\mathbf{e_{ij}}$ from the bilinear layer is $64$ followed by linear layers to project the edge features to edge probabilities. For baseline methods, we follow the same experimental settings discussed in~\cite{huang2020skipgnn}.

\section{Results}
In this section, we investigate the performance and flexibility of HOGCN on interaction prediction using four different datasets. We further explore the robustness of HOGCN to sparse networks. Finally, we demonstrate the ability of HOGCN to make novel predictions with literature-based case studies.

\subsection{Biomedical interaction prediction}
We compare HOGCN against various baselines on biomedical interaction prediction tasks using four different types of interaction datasets including protein-protein interactions (PPIs), drug-target interactions (DTIs), drug-drug interactions (DDIs) and gene-disease associations (GDIs). 

We randomly mask $20\%$ of interactions from the network as a test set and $10\%$ as a validation set. We train all models with $70\%$ of interactions and evaluate their performances on test sets. The best set of hyperparameters is selected based on their performances on the validation dataset. Finally, the experiment is repeated for five independent random splits of the interaction dataset and the results with $\pm$ one standard deviation are reported in Table~\ref{table_results}. All of our models used for the reported results are of same capacity (i.e. $P = \{0, 1, 2, 3\}$ and $d = 32$).

\begin{table}[htb]
\caption{Average AUPRC and AUROC with $\pm$ one standard deviation on biomedical interaction prediction}
\label{table_results}
    \centering
    \renewcommand{\arraystretch}{1.2}
    \begin{tabular}{l|ccc}
    \hline
Dataset & Method & AUPRC & AUROC \\
\hline
\multirow{9}{*}{DTI} & DeepWalk & 0.753 $\pm$ 0.008 & 0.735 $\pm$ 0.009 \\
& node2vec 		& 0.771 $\pm$ 0.005 	& 0.720 $\pm$ 0.010  \\
& L3			& 0.891 $\pm$ 0.004 	& 0.793 $\pm$ 0.006 \\
& VGAE			& 0.853 $\pm$ 0.010 	& 0.800 $\pm$ 0.010 \\
& GCN			& 0.904 $\pm$ 0.011 	& 0.899 $\pm$ 0.010 \\
& SkipGNN 		& 0.928 $\pm$ 0.006 	& 0.922 $\pm$ 0.004 \\
\cdashline{2-4}
& HOGCN & \textbf{0.937  $\pm$ 0.001} & \textbf{0.934 $\pm$ 0.001} \\
\hline
\multirow{9}{*}{DDI} & DeepWalk 	& 0.698 $\pm$ 0.012 & 0.712 $\pm$ 0.009 \\ 
& node2vec 	& 0.801 $\pm$ 0.004 & 0.809 $\pm$ 0.002 \\ 
& L3			& 0.860 $\pm$ 0.004 & 0.869 $\pm$ 0.003 \\ 
& VGAE		& 0.844 $\pm$ 0.076 & 0.878 $\pm$ 0.008 \\
& GCN			& 0.856 $\pm$ 0.005 & 0.875 $\pm$ 0.004 \\
& SkipGNN		& 0.866 $\pm$ 0.006 & 0.886 $\pm$ 0.003 \\
\cdashline{2-4}
& HOGCN & \textbf{0.897 $\pm$ 0.003} & \textbf{0.911 $\pm$ 0.002} \\
\hline
\multirow{9}{*}{PPI} & DeepWalk 	&0.715 $\pm$ 0.008 &0.706 $\pm$ 0.005\\ 
& node2vec 	&0.773 $\pm$ 0.010 &0.766 $\pm$ 0.005\\ 
& L3			&0.899 $\pm$ 0.003 &0.861 $\pm$ 0.003\\
& VGAE		&0.875 $\pm$ 0.004 &0.844 $\pm$ 0.006\\
& GCN			&0.909 $\pm$ 0.002 &0.907 $\pm$ 0.006\\
& SkipGNN 	&   0.921 $\pm$ 0.003 & 0.917 $\pm$ 0.004\\
\cdashline{2-4}
& HOGCN & \textbf{0.930 $\pm$ 0.002} & \textbf{0.922 $\pm$ 0.001} \\
\hline 
\multirow{9}{*}{GDI} & DeepWalk 		& 0.827 $\pm$ 0.007 & 0.832 $\pm$ 0.003  \\
& node2vec 		& 0.828 $\pm$ 0.006 & 0.834 $\pm$ 0.003  \\
& L3			& 0.899 $\pm$ 0.001 & 0.832 $\pm$ 0.001 \\
& VGAE			& 0.902 $\pm$ 0.006 & 0.873 $\pm$ 0.009 \\
& GCN			& 0.909 $\pm$ 0.002 & 0.906 $\pm$ 0.006 \\
& SkipGNN 		& 0.915 $\pm$ 0.003 & 0.912 $\pm$ 0.004 \\
\cdashline{2-4}
& HOGCN  & \textbf{0.941 $\pm$ 0.001} & \textbf{0.936 $\pm$ 0.001} \\
\hline
\end{tabular}
\end{table}

Table~\ref{table_results} shows that HOGCN achieves huge improvement over network embedding methods such as DeepWalk and node2vec across all datasets. Specifically, HOGCN outperforms Deepwalk on AUPRC by 24.44\% in DTI, 28.51\% in DDI, 30.07\% in PPI, and 13.79\% in GDI. Although node2vec achieves better performance compared to DeepWalk by adopting a biased random walk, HOGCN still outperforms node2vec by a significant margin. DeepWalk and node2vec consider different orders of neighborhood defined by the window size and learns similar representations for the nodes in that window. In contrast, HOGCN learns feature differences between neighbors at various distances to obtain feature representation for the node and thus achieves superior performance. The improved performance suggests that feature differences between different order neighbors provide important information for interaction prediction.

A network similarity-based method, L3~\cite{kovacs2019network} outperforms network embedding methods across four datasets but is limited to a single aspect of network similarity i.e. the number of paths of length $3$ connecting two nodes. So, L3 cannot be applied when other similarities between nodes such as similarity in features and common neighbors at various distances need to be considered. HOGCN overcomes these limitations and outperforms L3 across all interaction datasets with huge gain. In particular, HOGCN gains 3.5\% AUPRC and 7.09\% AUROC on PPI over L3~\cite{kovacs2019network}, which recently outperformed 20 network science methods in the PPI prediction problem.

\begin{figure*}[htb]
\centering
\subfloat[]{\includegraphics[width=0.32\textwidth]{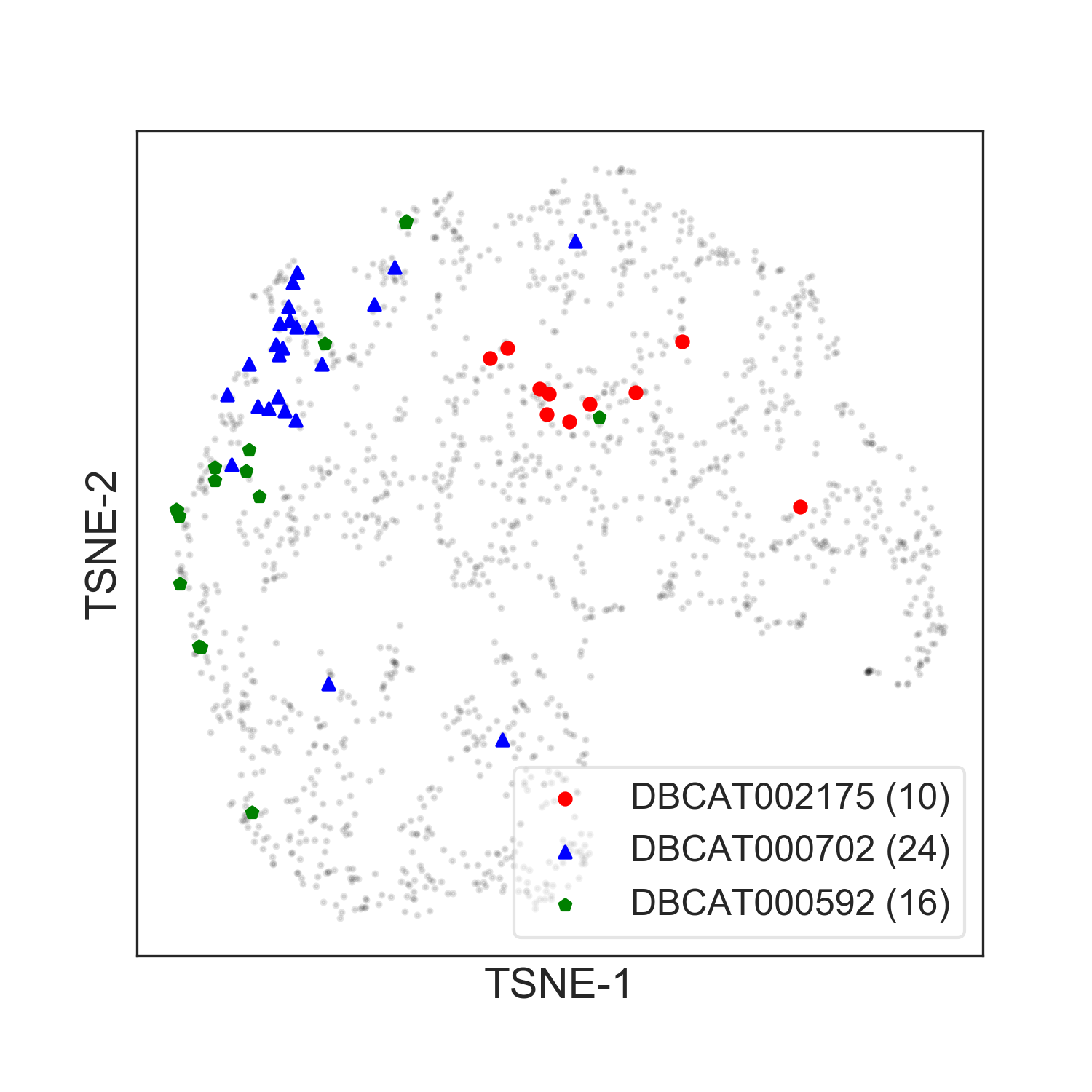}%
\label{fig_emb_1}}
\hfil
\subfloat[]{\includegraphics[width=0.32\textwidth]{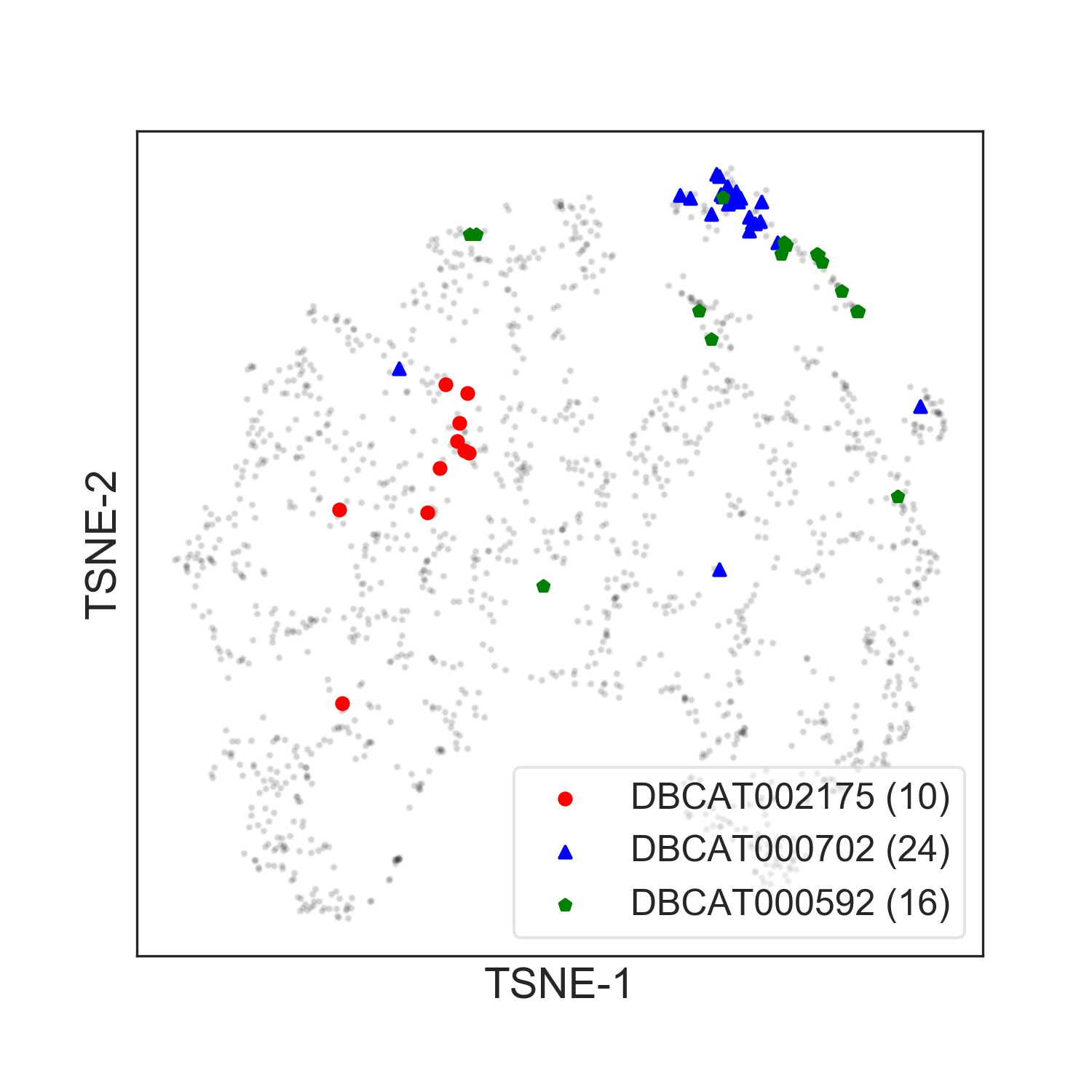}%
\label{fig_emb_2}}
\hfil
\subfloat[]{\includegraphics[width=0.32\textwidth]{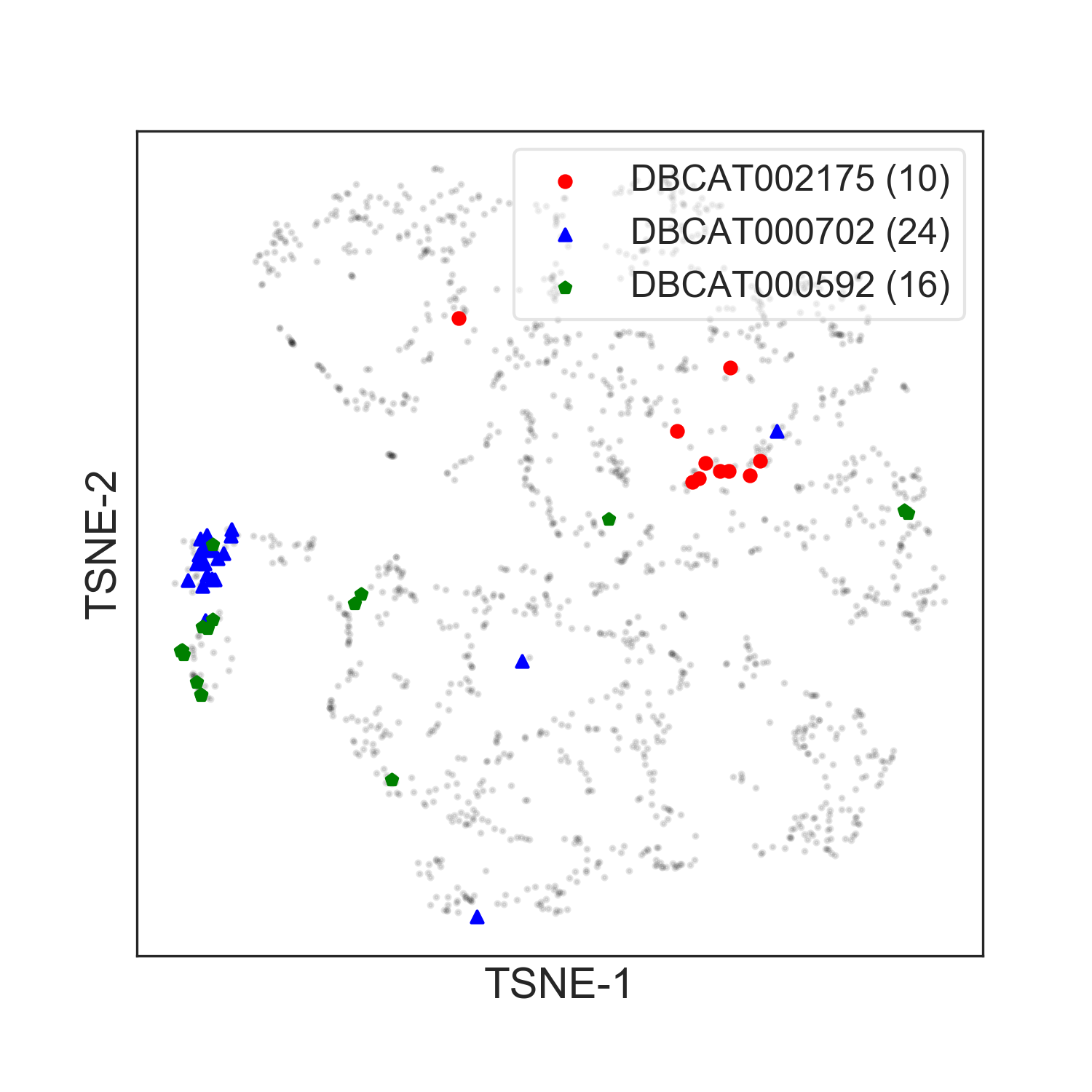}%
\label{fig_emb_3}}
\caption{ Visualization of learned representation for drugs with (a) GCN (b) SkipGNN (c) HOGCN. Drugs are mapped to the 2D space using t-SNE package~\cite{maaten2008visualizing} with learned drug representations. Drugs categories such as DBCAT002175, DBCAT000702 and DBCAT000592 are highlighted. The number of drugs in each categroy is reported in legend. Best viewed on screen.}
\label{fig_drug_embeddings}
\end{figure*}

Graph convolution-based methods such as GCN and VGAE achieves significant improvements over network embedding approaches but achieves comparable performance with L3. SkipGNN shows improvement over all other methods by incorporating skip similarity to aggregate information from second-order neighbors. Moreover, HOGCN with $k=3$ achieves an improvement over all graph convolution-based methods. Specifically, HOGCN achieves improvement in AUPRC over VGAE~\cite{kipf2016variational} by 6.3\%, GCN~\cite{kipf2016semi} by 4.8\% and SkipGNN~\cite{huang2020skipgnn} by 3.6\% on DDI dataset. 

As HOGCN can learn the linear combination of node features at multiple distances, it can extract meaningful representations from the interaction networks. The results in Table~\ref{table_results} demonstrate that our approach with higher-order neighborhood mixing outperforms the state-of-the-art methods on real interaction datasets.

\subsection{Exploration of HOGCN's drug representations}
Next, we evaluate if HOGCN learns meaningful representation when feature representations of higher-order neighbors are aggregated. To this aim, we train GCN, SkipGNN, and HOGCN models on the DDI network to obtain the drug representations $Z$. The learned drug representations are mapped to 2D space using t-SNE~\cite{maaten2008visualizing} and visualize them in Fig.~\ref{fig_drug_embeddings}.

Drugbank~\cite{wishart2006drugbank} provides information about drugs and their categories based on different characteristics such as involved metabolic enzymes, class of drugs, side effects of drugs, and the like. For this experiment, we collect drug categories from Drugbank and limit the selection of drug categories such that the training set doesn't contain any interactions between the drugs in the same category. The selected drug categories are ACE Inhibitors and Diuretics (DBCAT002175), Penicillins (DBCAT000702), and Antineoplastic Agents (DBCAT000592) with 10, 24, and 16 drugs respectively. Although these drugs don't have direct interactions in the training set, we assume that these drugs share neighborhoods at various distances and can be explored accordingly with HOGCN. 

Fig.~\ref{fig_drug_embeddings} shows the clustering structure in drugs' representations as neighborhood information at multiple distances are considered. Examining the figure, we observe that drugs in the same category are embedded close to each other in the 2D space when the model aggregates information from farther neighbors. For example, 24 drugs in the Penicillins (DBCAT000702) category (marked with blue triangles in Fig.~\ref{fig_drug_embeddings}) are scattered in the representation space learned by GCN that only considers feature aggregation from immediate neighbors (Fig.~\ref{fig_emb_1}). Note that these drugs don't have any direct interaction between themselves in the training set. Since GCN-based models can only average the representation from immediate neighbors, these drugs are mapped relatively farther to each other and closer to other interacting drugs. SkipGNN considers skip similarity to aggregate features from second-order neighbors and show relatively compact clusters compared to GCNs (Fig.~\ref{fig_emb_2}). On the other hand, HOGCN considers the higher-order neighborhood and learns similar representations for drugs that belong to the same category demonstrated by compact clustering structure in Fig.~\ref{fig_emb_3} even though no information about categorical similarity is provided to the model. This analysis demonstrates that HOGCN learns meaningful representation for drugs by aggregating feature representations from the neighborhood at various distances.

Next, we test if the clustering pattern in Fig.~\ref{fig_drug_embeddings} holds across many drug categories. With this aim, we consider all drug categories in DrugBank and compute the average Euclidean distance between each drug's representation and representations of other drugs within the same drug category. We then perform 2-sample Kolmogorov–Smirnov tests and found that HOGCN learns significantly more similar representations of drugs than expected by chance (p-value = $4.93e-106$), GCNs (p-value = $5.05e-56$) and SkipGNN ($1.29e-12$). Thus, this analysis indicates that HOGCN learns meaningful representations for drugs by aggregating neighborhood information at various distances.

\subsection{Robustness to network sparsity}\label{network_sparsity}
We next explore the robustness of network-based interaction prediction models to network sparsity. To this aim, we evaluate the performance with respect to the percentage of training edges varying from 10\% to 70\%. We make predictions on the rest of the interactions. We further use 10\% of test edges for validation to select the best hyperparameter settings. For a fair comparison, we compare with graph convolution-based methods that aggregate information from direct and/or second-order neighbors.

\begin{figure}[htb]
\centering
\subfloat[]{\includegraphics[width=0.49\linewidth]{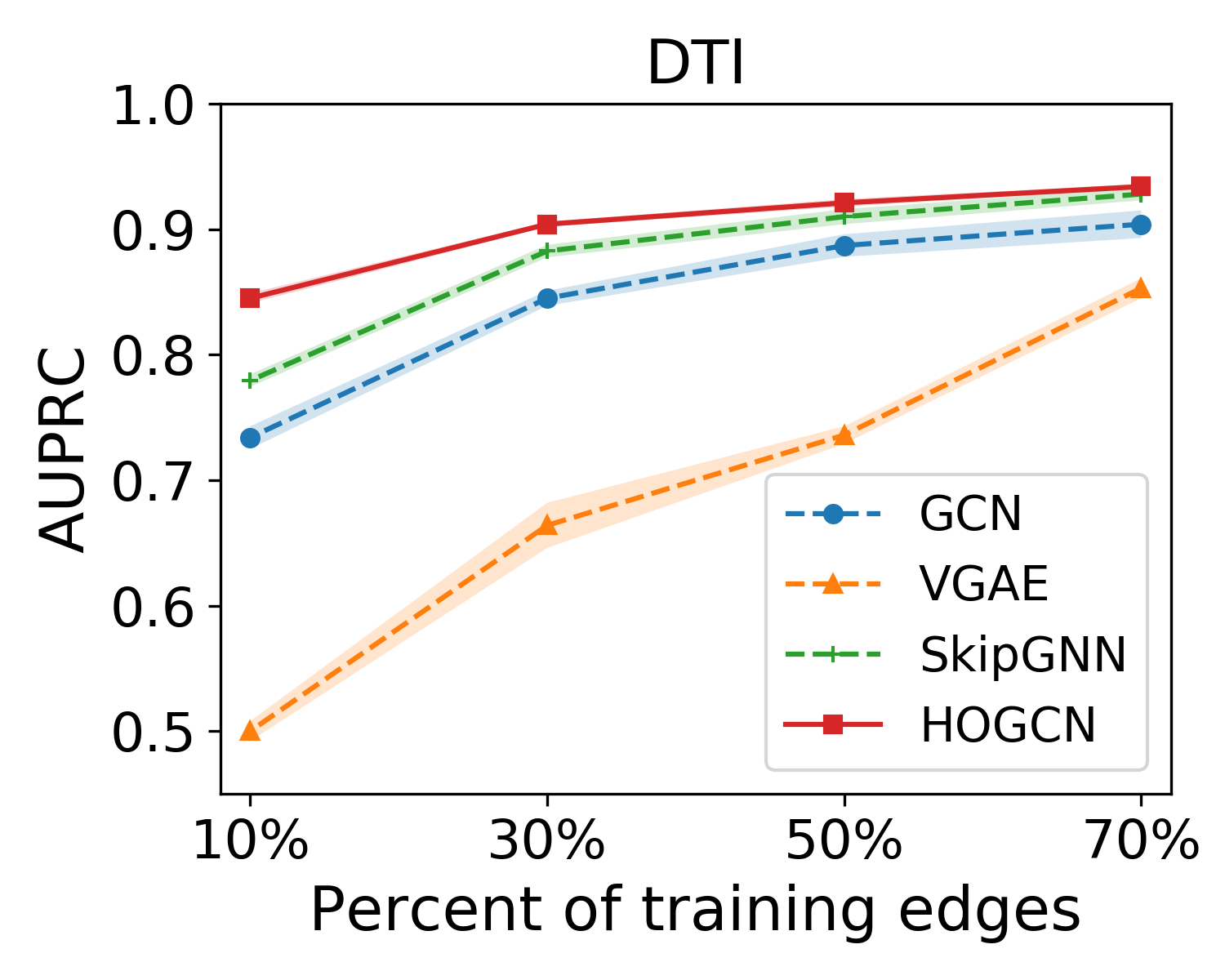}%
\label{fig_dti}}
\hfil
\subfloat[]{\includegraphics[width=0.49\linewidth]{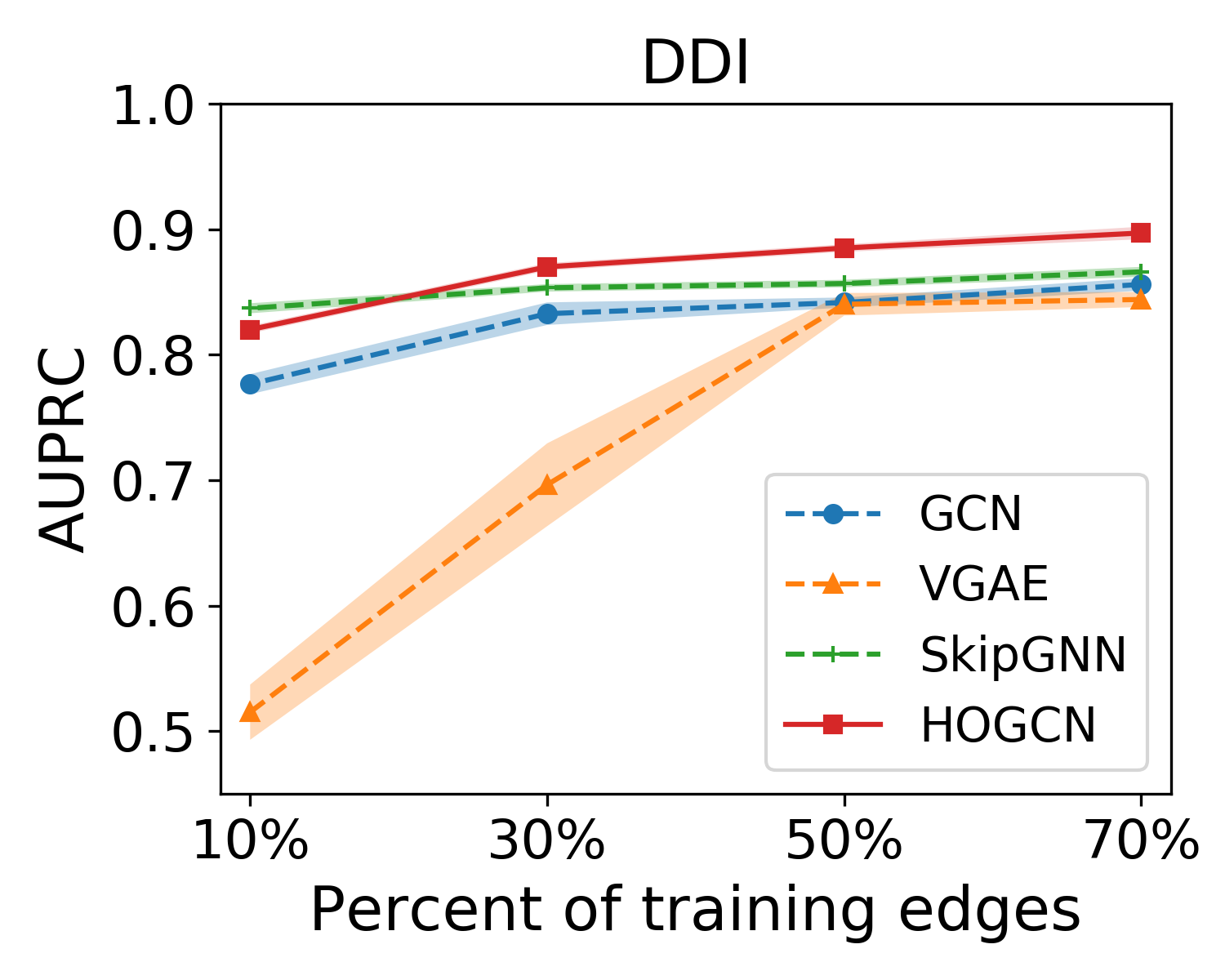}%
\label{fig_ddi}}
\hfil
\subfloat[]{\includegraphics[width=0.49\linewidth]{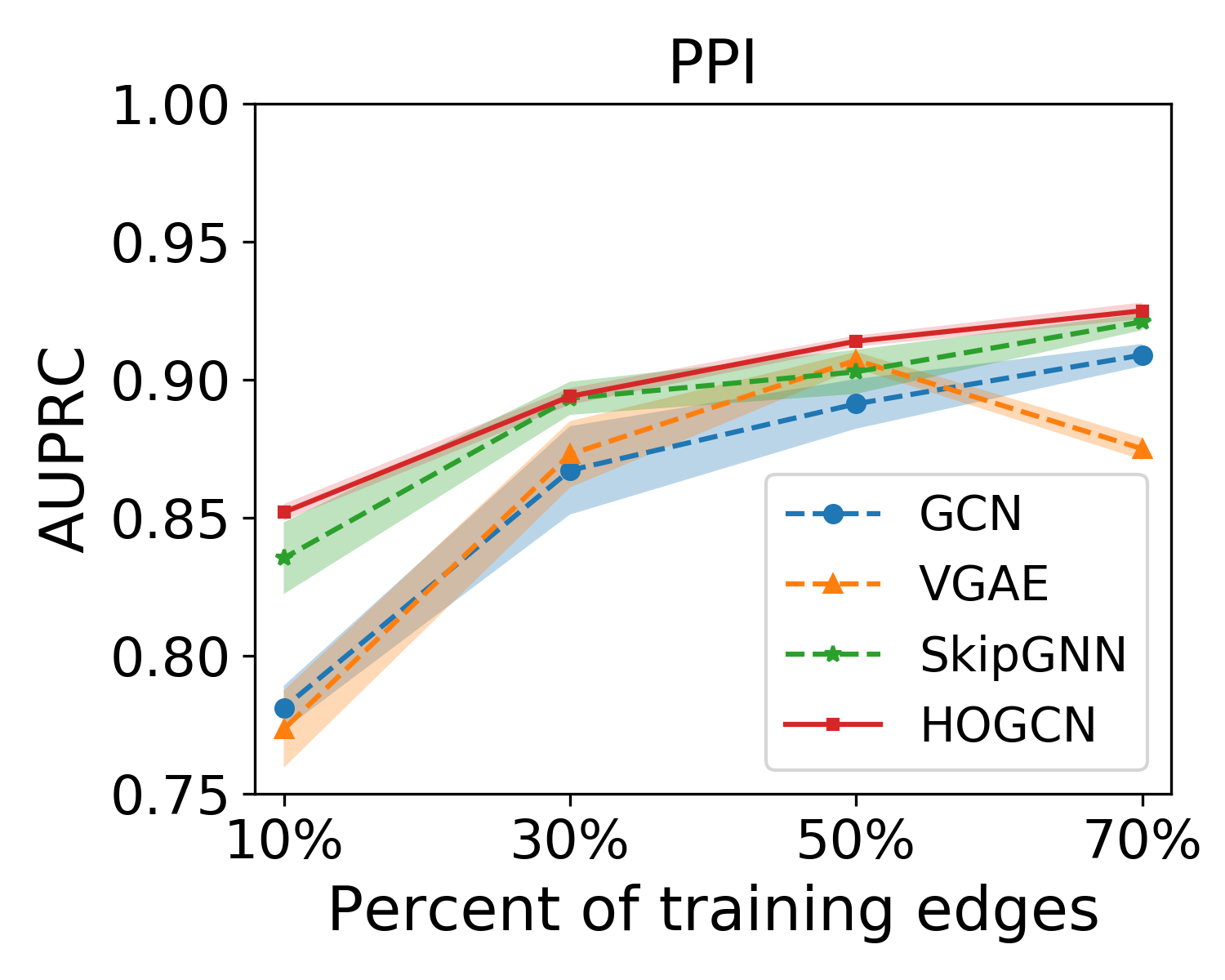}%
\label{fig_ppi}}
\subfloat[]{\includegraphics[width=0.49\linewidth]{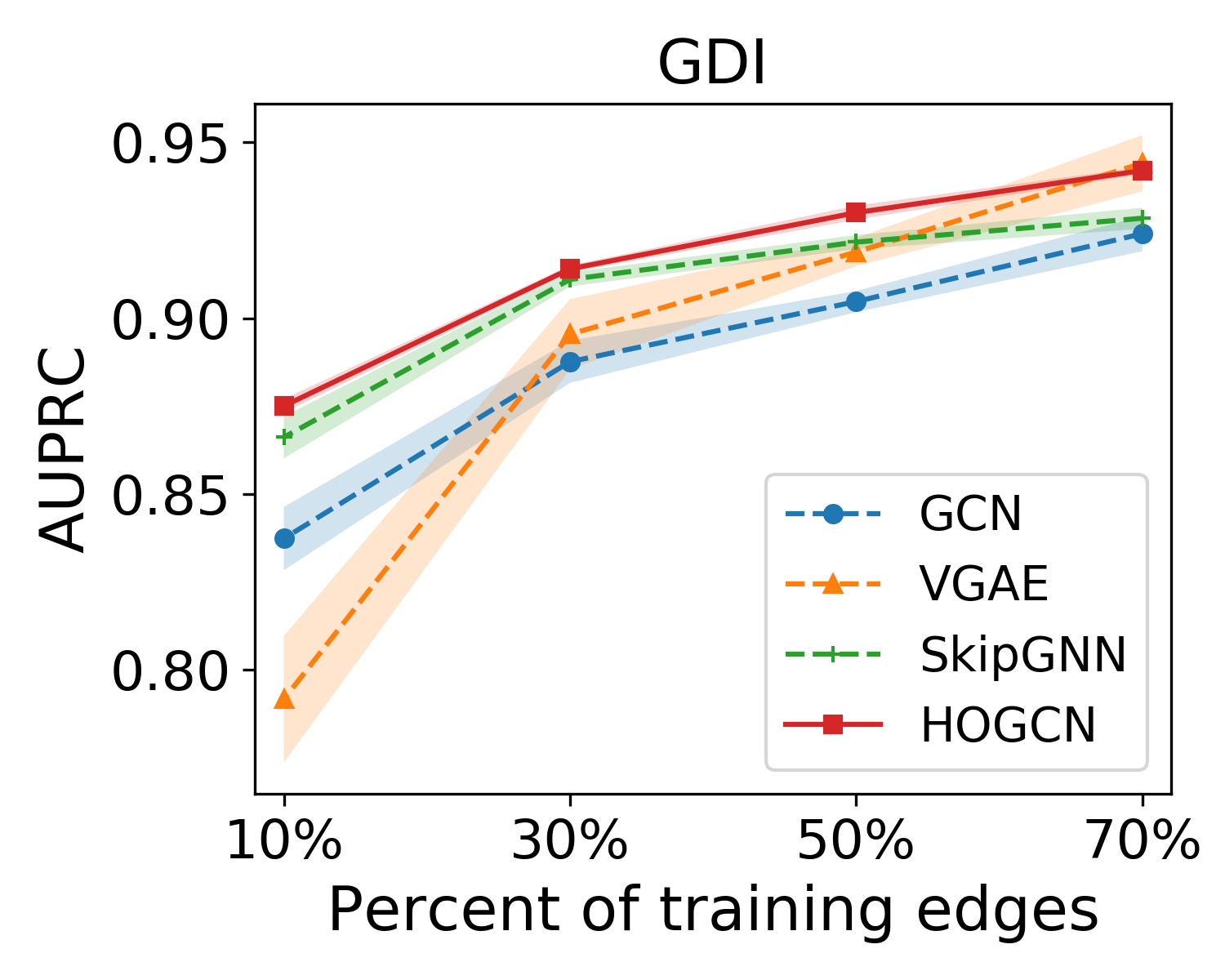}%
\label{fig_gdi}}
\caption{AUPRC comparison of HOGCN's performance with that of alternative approaches with respect to network sparsity. HOGCN consistently achieves better performance in various fraction of training edges.}
\label{fig_training_ratios}
\end{figure}

Fig.~\ref{fig_training_ratios} shows the robustness of HOGCN to network sparsity. HOGCN achieves strong performance in all tasks with different network sparsity. The performance of HOGCN steadily improves with the increase in training edges. The mixing of features from a higher-order neighborhood in HOGCN and SkipGNN shows improvement over GCN and VGAE that only consider direct neighbors. Since HOGCN can learn the linear combination of features from a 3-hop neighborhood for this experiment, it shows improvement over SkipGNN in almost all cases. This demonstrates that features from farther distances are informative for interaction prediction in sparse networks.

\subsection{Calibrating model's prediction}\label{calibration}
All graph convolution-based model proposed for biomedical link prediction predicts the confidence estimate $p_{ij}$ for interaction between two biomedical entities $v_i$ and $v_j$. We thus test if a predicted confidence $p_{ij}$ represents the likelihood of being true interaction. In other words, we expect the confidence estimate $p_{ij}$ to be calibrated, \textit{i.e.} $p_{ij}$ represents true interaction probability~\cite{niculescu2005predicting}. For example, given 100 predictions, each with the confidence of 0.9, we expect that 90 interactions should be correctly classified as true interactions.

\begin{figure}[htb]
\centering
\subfloat[]{\includegraphics[width=0.49\linewidth]{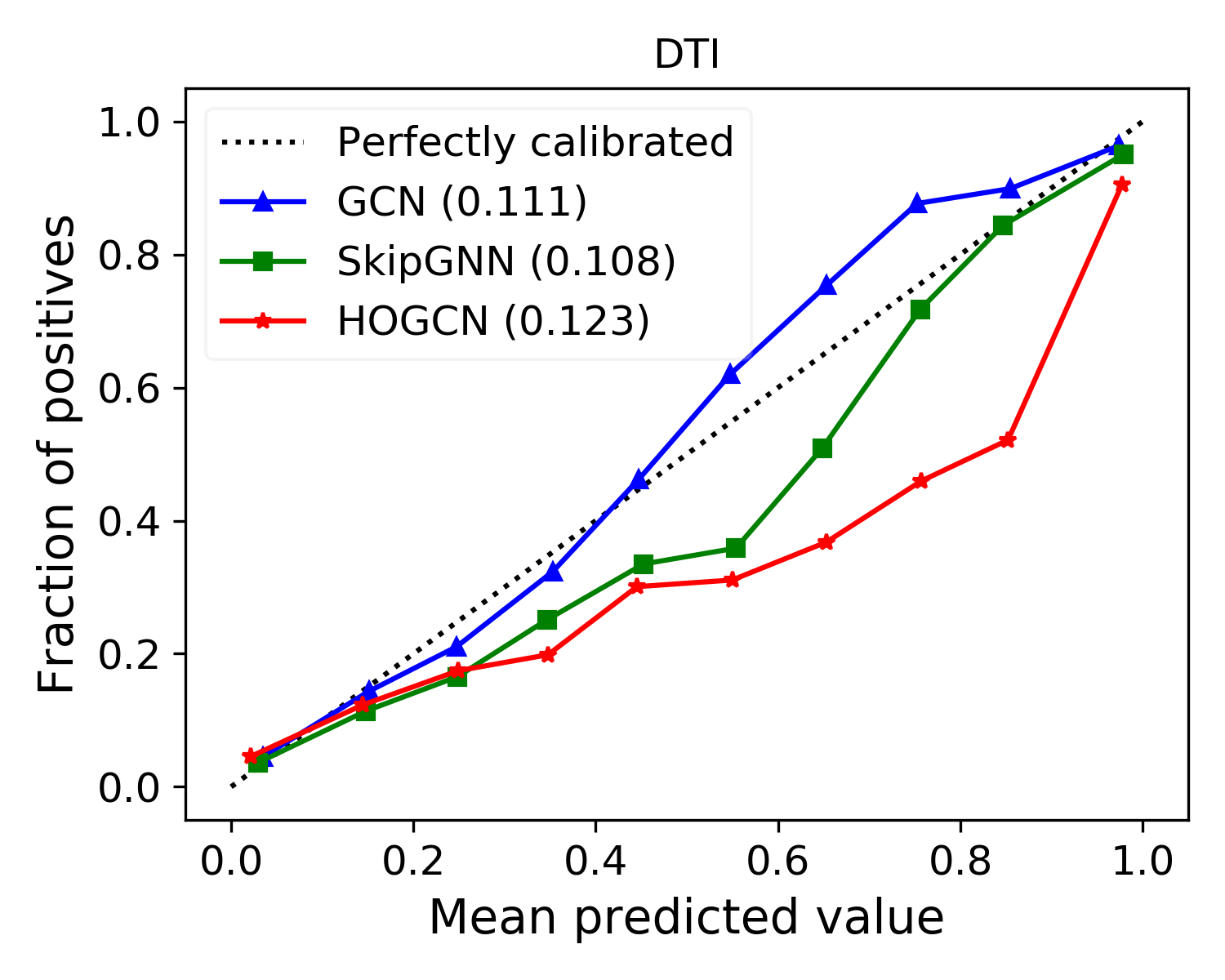}%
\label{fig_dti_calibration}}
\hfil
\subfloat[]{\includegraphics[width=0.49\linewidth]{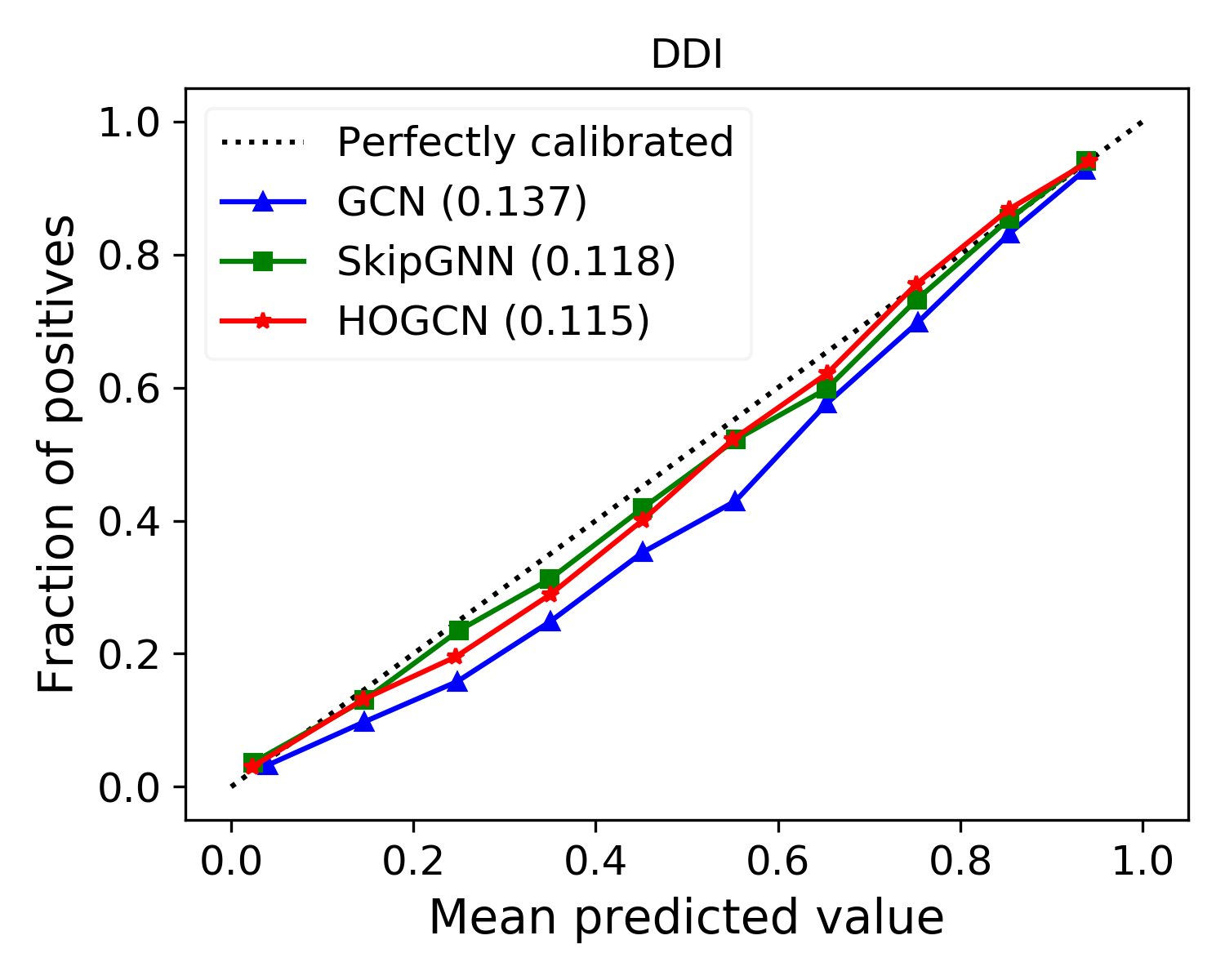}%
\label{fig_ddi_calibration}}
\hfil
\subfloat[]{\includegraphics[width=0.49\linewidth]{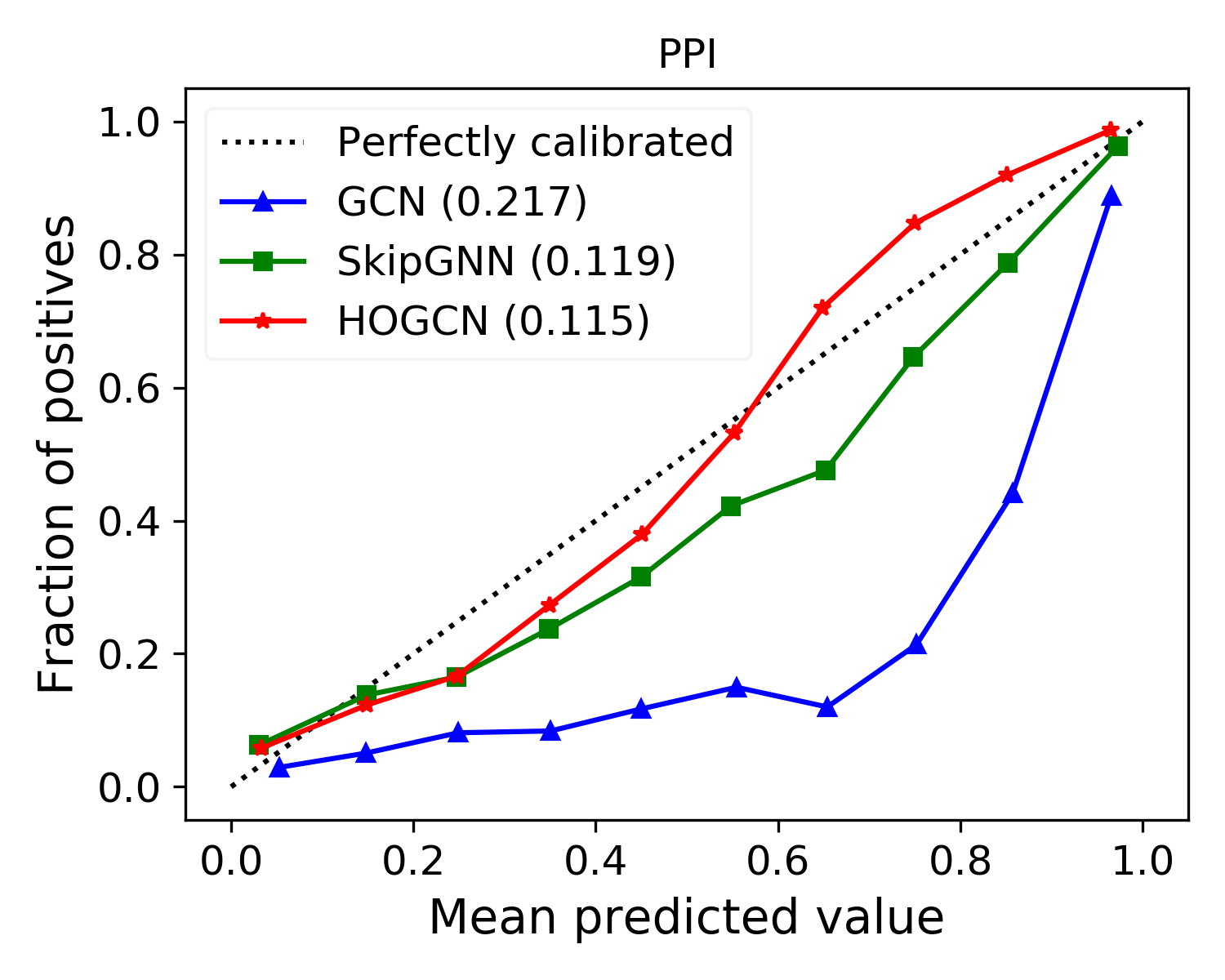}%
\label{fig_ppi_calibration}}
\subfloat[]{\includegraphics[width=0.49\linewidth]{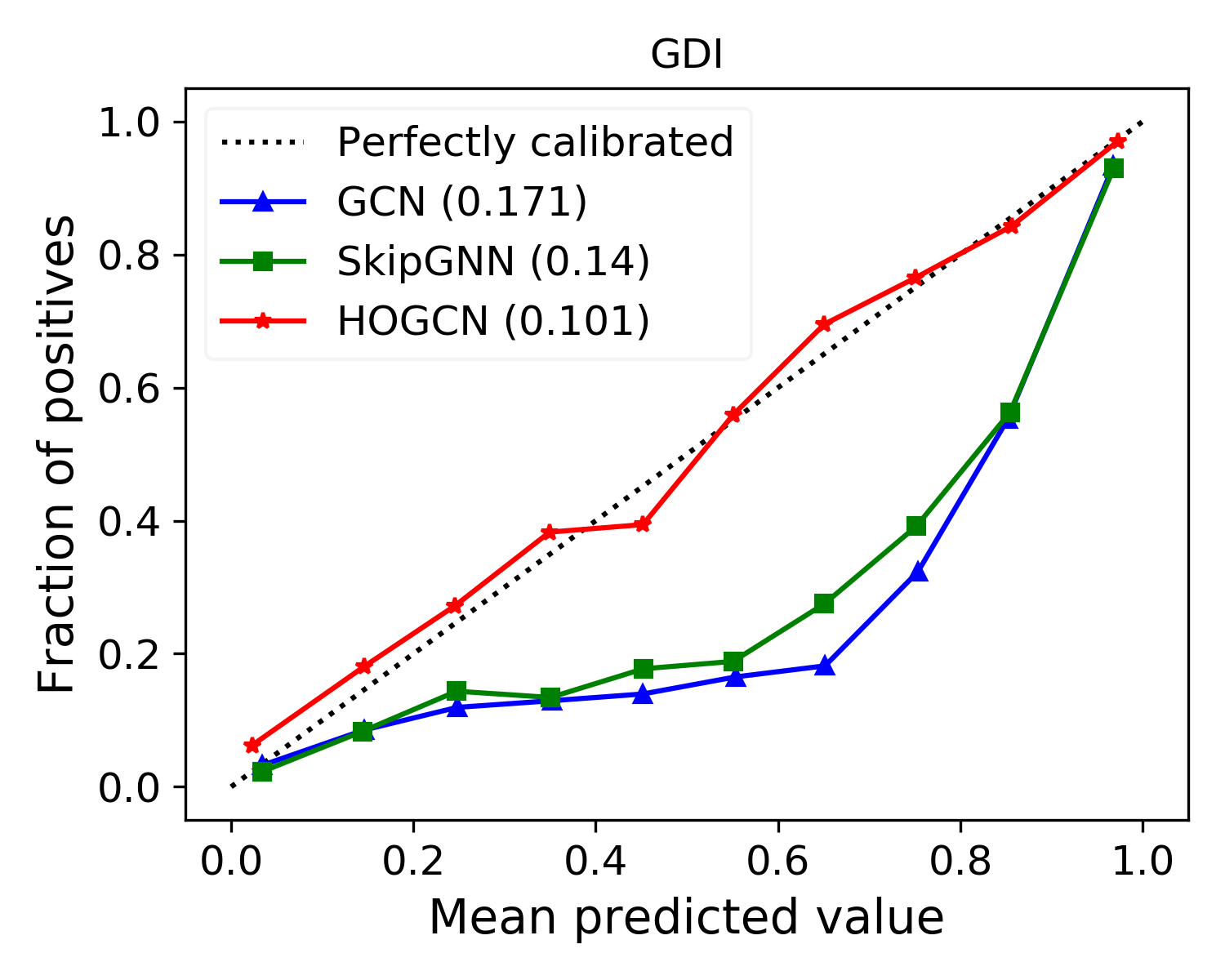}%
\label{fig_gdi_calibration}}
\caption{Reliability diagrams for different graph convolution-based methods. The calibration performance is evaluated with Brier score, reported in the legend (lower is better).}
\label{fig_calibration}
\end{figure}

To evaluate the calibration performance, we use reliability diagrams~\cite{niculescu2005predicting} and Brier score~\cite{brier1950verification}. In particular, the reliability diagram provides a visual representation of model calibration. These diagrams plot the expected fraction of positives as a function of predicted probability~\cite{niculescu2005predicting}. A model with perfectly calibrated predictions is represented by a diagonal in Fig.~\ref{fig_calibration}. In addition to reliability diagrams, it is more convenient to have a scalar summary statistics of calibration. Brier score~\cite{brier1950verification} is a proper scoring rule for measuring the accuracy of predicted probabilities. Lower Brier score indicates better calibration of a set of predictions. It is computed as the mean squared error of a predicted probability $p_{ij}$ and the ground-truth interaction label $A_{ij}$. Mathematically, the Brier score can be computed as:
\begin{equation}
\text{Brier score} = \frac{1}{|\mathcal{E}^{\prime}|}\sum\limits _{(i,j)=1}^{|\mathcal{E}^{\prime}|}(p_{ij}-A_{ij})^2    
\end{equation}
where $|\mathcal{E}^{\prime}|$ denotes the number of test edges.

Fig.~\ref{fig_calibration} shows the calibration plots for GCN, SkipGNN and HOGCN ($k=3$). For DTI dataset, SkipGNN show better calibration compared to GCN and HOGCN (Fig.~\ref{fig_dti_calibration}), indicating that second-order neighborhood information is appropriate and aggregating features from farther away makes model overconfident. For other datasets, GCNs are relatively overconfident for all predicted confidence. For example, approximately $20\%-30\%$ of interactions are true positives among the interactions with high predicted confidence $0.8$ in PPI (Fig.~\ref{fig_ppi_calibration}) and GDI dataset (Fig.~\ref{fig_gdi_calibration}). In contrast, HOGCN achieves a lower Brier score in comparison to the GCN and SkipGNN across DDI, PPI, and GDI datasets, alluding to the benefits of aggregating higher-order neighborhood features for calibrated prediction. This analysis demonstrates that HOGCN with higher-order neighborhood mixing makes accurate and calibrated predictions for biomedical interaction.

\subsection{Impact of higher-order neighborhood mixing}
In Section~\ref{network_sparsity}, we contrast HOGCN's performance with that of alternative graph convolution-based methods in varying fraction of edges. In this experiment, we aim to observe the performance of HOGCN when the order $k$ is increased to allow the model to aggregate neighborhood information from farther away. We follow a similar setup as discussed in~\ref{network_sparsity}.

\begin{figure}[htb]
\centering
\subfloat[]{\includegraphics[width=0.49\linewidth]{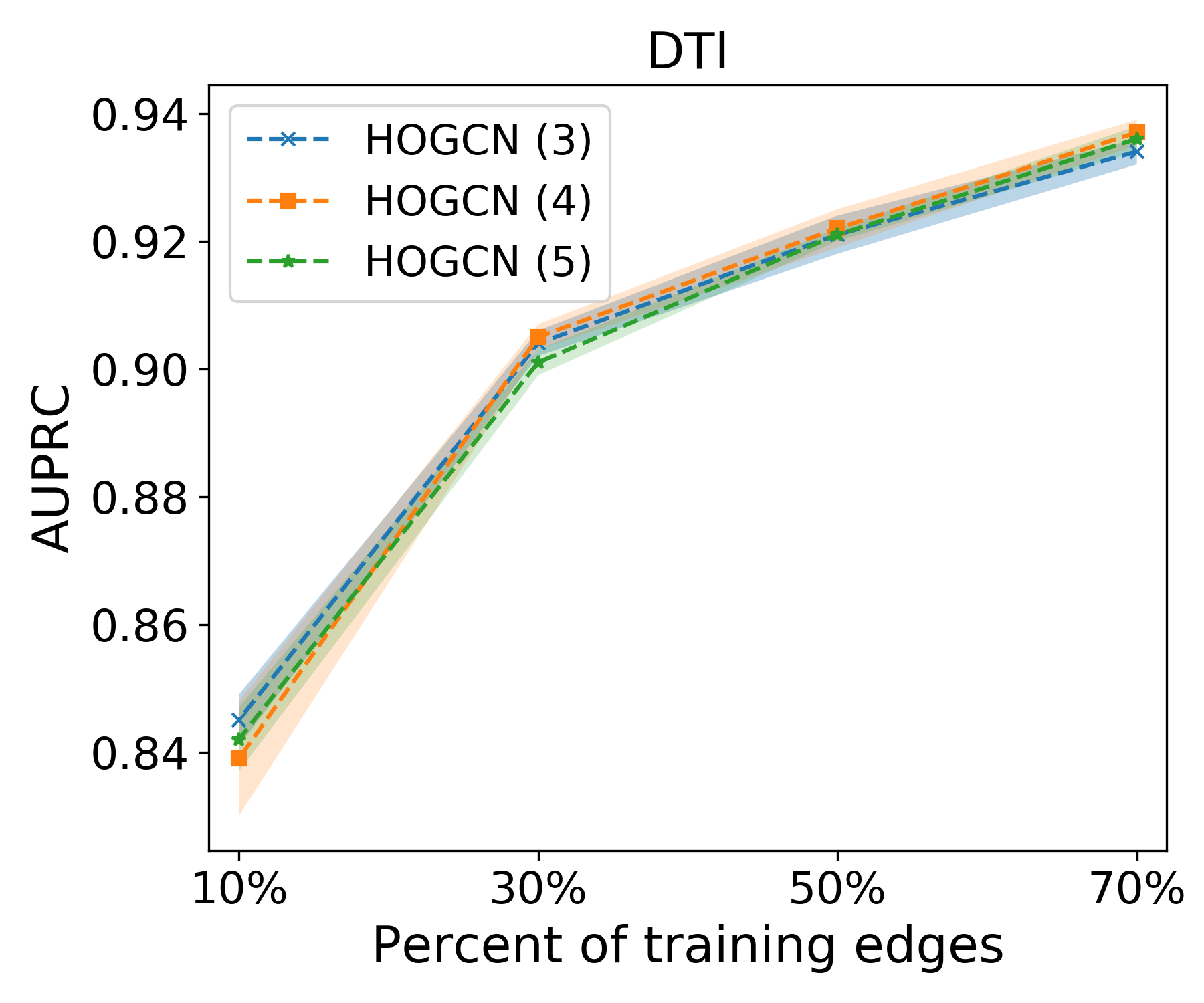}%
\label{fig_dti_order}}
\hfil
\subfloat[]{\includegraphics[width=0.49\linewidth]{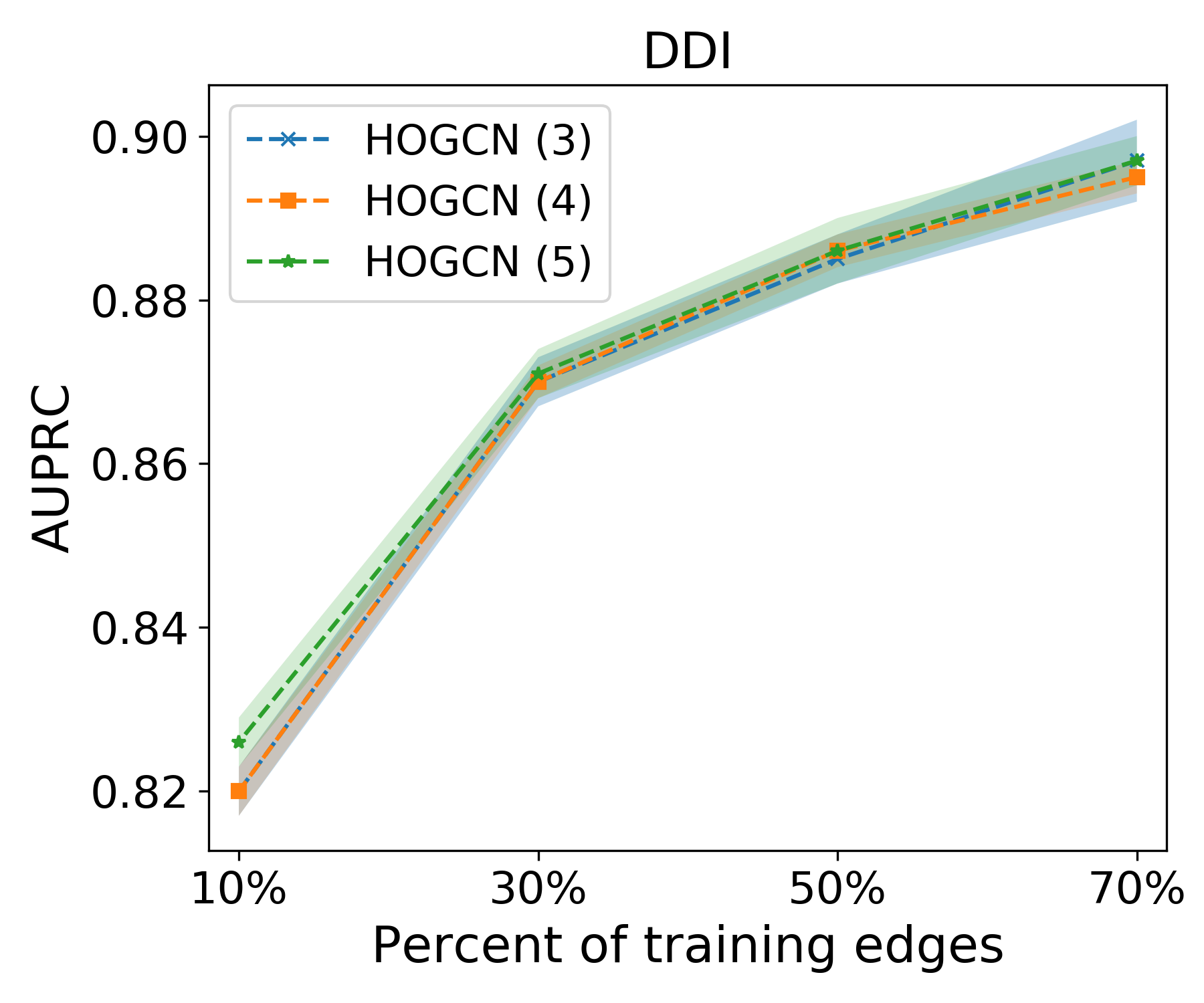}%
\label{fig_ddi_order}}
\hfil
\subfloat[]{\includegraphics[width=0.49\linewidth]{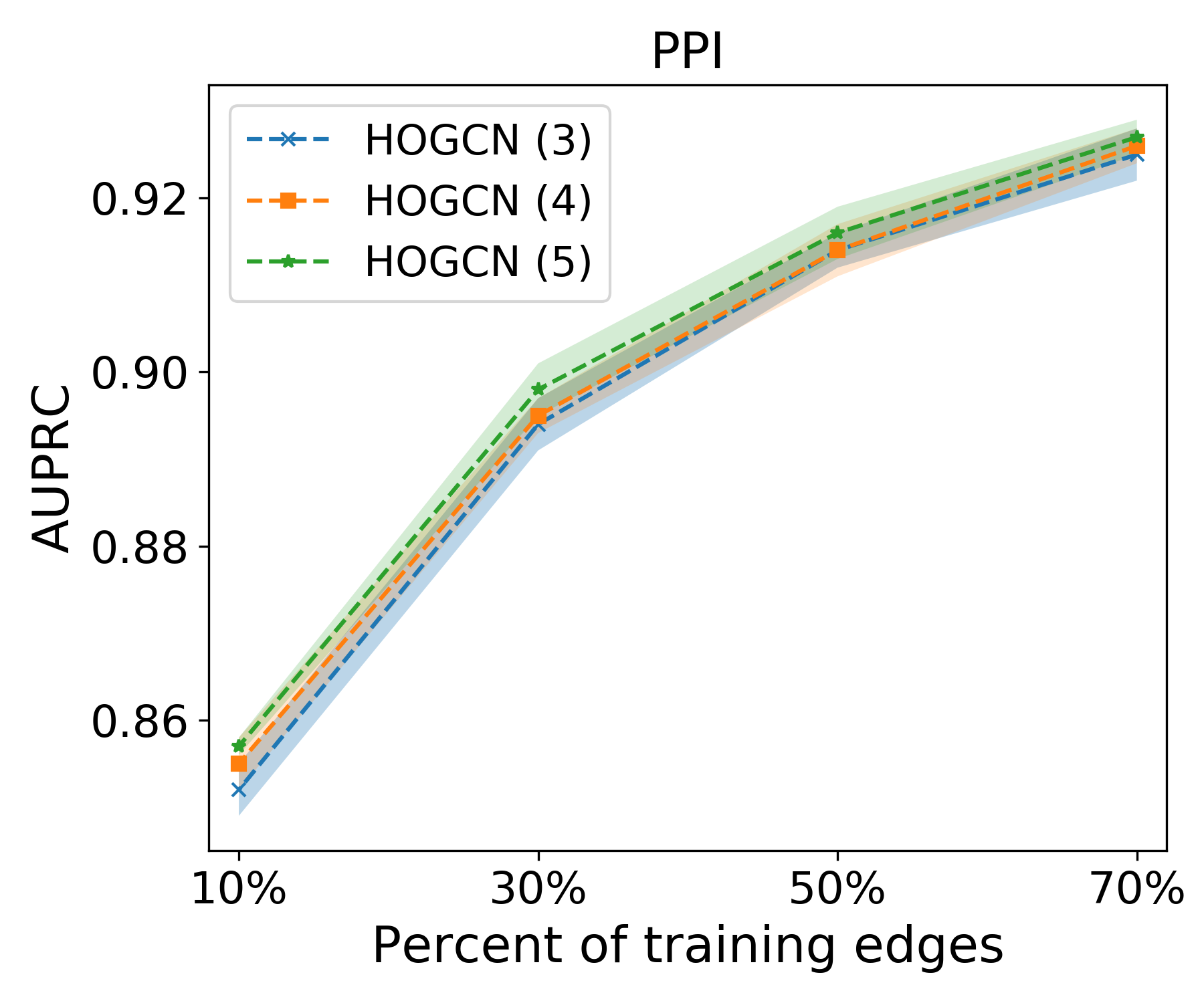}%
\label{fig_ppi_order}}
\subfloat[]{\includegraphics[width=0.49\linewidth]{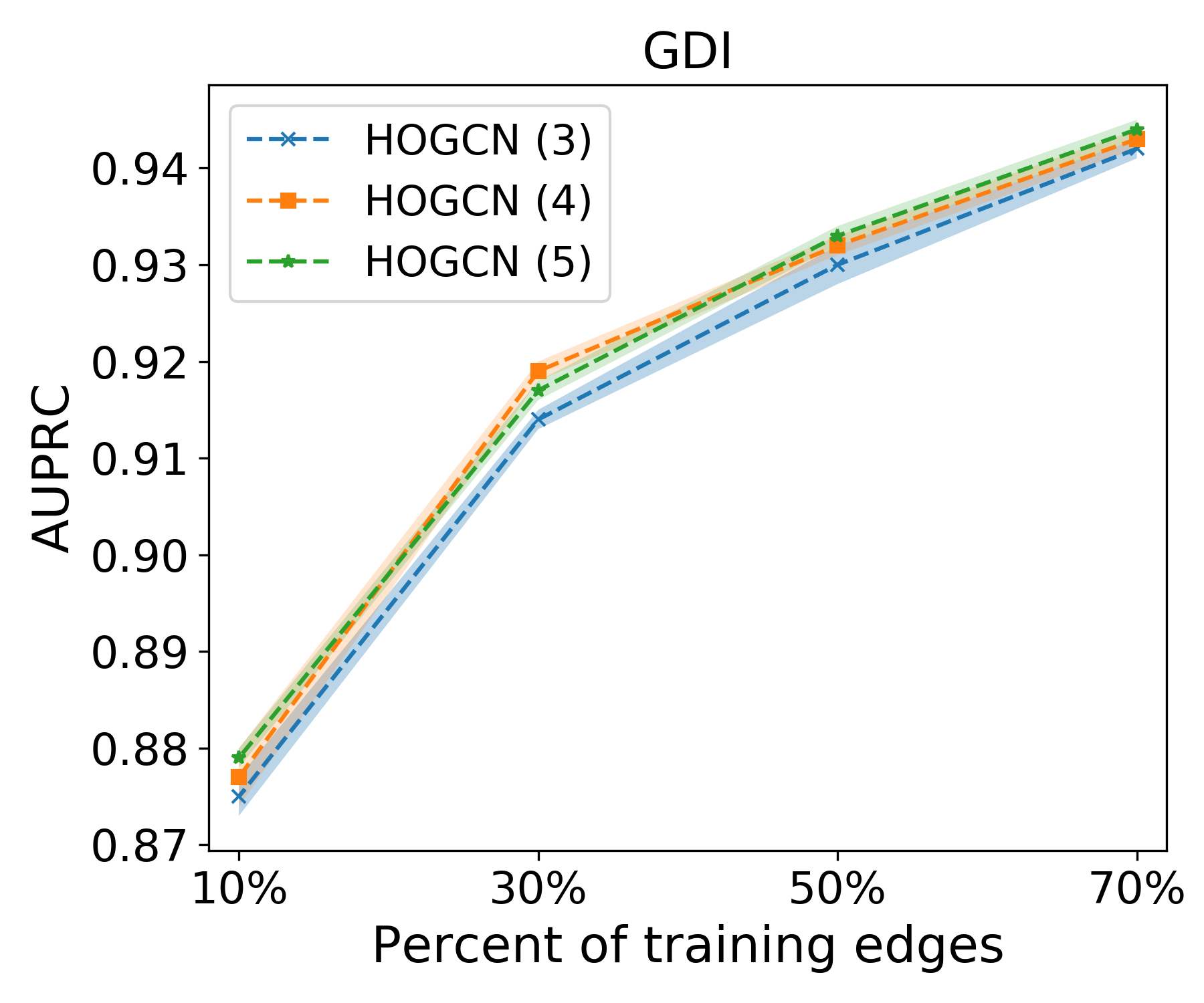}%
\label{fig_gdi_order}}
\caption{AUPRC comparison for higher-order message passing with different fractions of training edges. The values of $k$ for different HOGCN models are reported in the legend.}
\label{fig_orders}
\end{figure}

Fig.~\ref{fig_orders} shows the  comparison of HOGCN with higher-order neighborhood mixing $k = \{3, 4, 5\}$. The prediction performance of HOGCN improves with the increase in the number of training interactions for all cases.  The results show that HOGCN’s performances are not sensitive to the hyperparameter settings of $k$ for all datasets since for settings $k=\{3, 4, 5\}$, we achieve comparable performances across the datasets. This analysis indicates that the 3-hop neighborhood provides sufficient information for interaction prediction across all datasets and the performance remains stable even with a large value for $k$.

\subsection{Investigation of novel predictions}
Next, we perform the literature-based validation of novel predictions. Our goal is to evaluate the quality of HOGCN's novel predictions compared to that of GCN and SkipGNN and show that HOGCN predicts novel interactions with higher confidence. We consider GDIs and DDIs for this evaluation.

We first evaluate the potential of the HOGCN to make novel GDI predictions. We collect 1,134,942 GDIs and their scores from DisGeNET~\cite{pinero2020disgenet}. The score corresponds to the number and types of sources and the number of publications supporting the associations. With the score threshold of $0.18$, we obtain 17,893 new GDIs that are not in the training set. We make predictions on these 17,893 GDIs with GCN, SkipGNN, and HOGCN ($k=3$). Out of 17,893 GDIs, HOGCN predicts a higher probability than GCN for 17,356 (96.99\%) GDIs and than SkipGNN for 11,418 (63.8\%) GDIs. Table~\ref{tab:gene_disease} shows the top 5 GDIs with a significant increase in interaction probabilities when higher-order neighborhood mixing is considered. We also provide the number of evidence from DisGenNet~\cite{pinero2020disgenet} to support these predictions. Improvement in predicted probabilities by HOGCN models shows that aggregating feature representations from higher-order neighbors make HOGCN more confident about the potential interactions as discussed in Section~\ref{calibration}.

\begin{table}[!htb]
\centering
\caption{Novel prediction of GDIs with the number of evidence from DisGenNet~\cite{pinero2020disgenet} supporting the interaction. GCN, SkipGNN and HOGCN are denoted by $1, 2$ and $3$ respectively.}
\label{tab:gene_disease}
\begin{tabular}{p{0.8cm}p{2.8cm}p{0.5cm}p{0.5cm}p{0.5cm}c}
\hline
&  & \multicolumn{3}{c}{Probability} & No. of \\
\cline{3-5}
Gene & Disease & $1$ & $2$ & $3$  & Evidence \\
\hline
PTGER1  & Gastric ulcer          & 0.087    & 0.519    &0.721   & 1\\
ANGPT1  & Gastric ulcer          & 0.173    & 0.583    &0.657   & 2\\
ABO	& Pancreatic carcinoma	& 0.265	& 0.615	& 0.770 & 26\\
VCAM1 &	Endotoxemia	& 0.294 & 0.529 & 0.639 & 5\\
GPC3    & Hepatoblastoma         & 0.307    & 0.540    &0.598   & 17\\
\hline
\end{tabular}
\end{table}

\begin{figure*}[htb]
\centering
\subfloat[]{\includegraphics[width=0.49\textwidth]{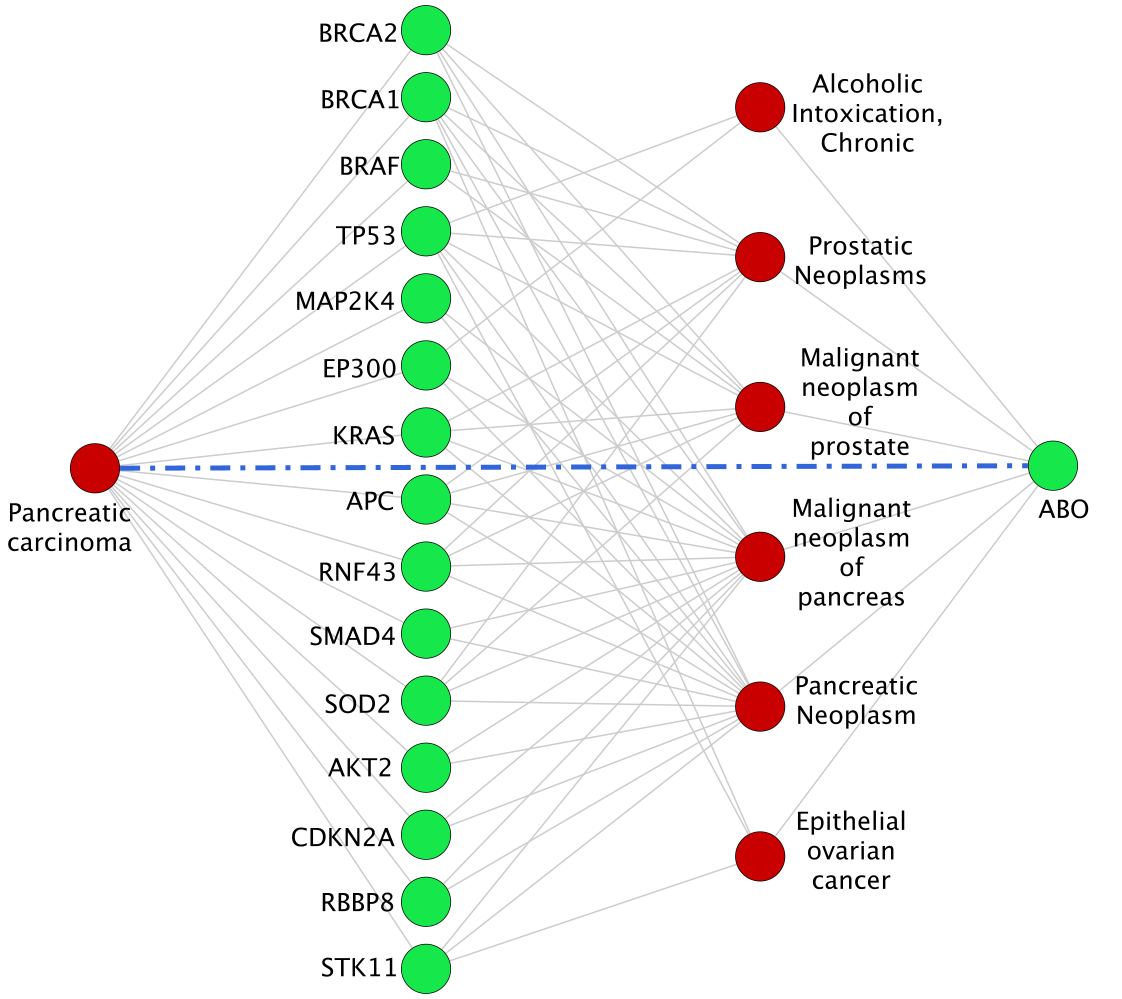}%
\label{fig_abo_pancreas}}
\hfil
\subfloat[]{\includegraphics[width=0.49\textwidth]{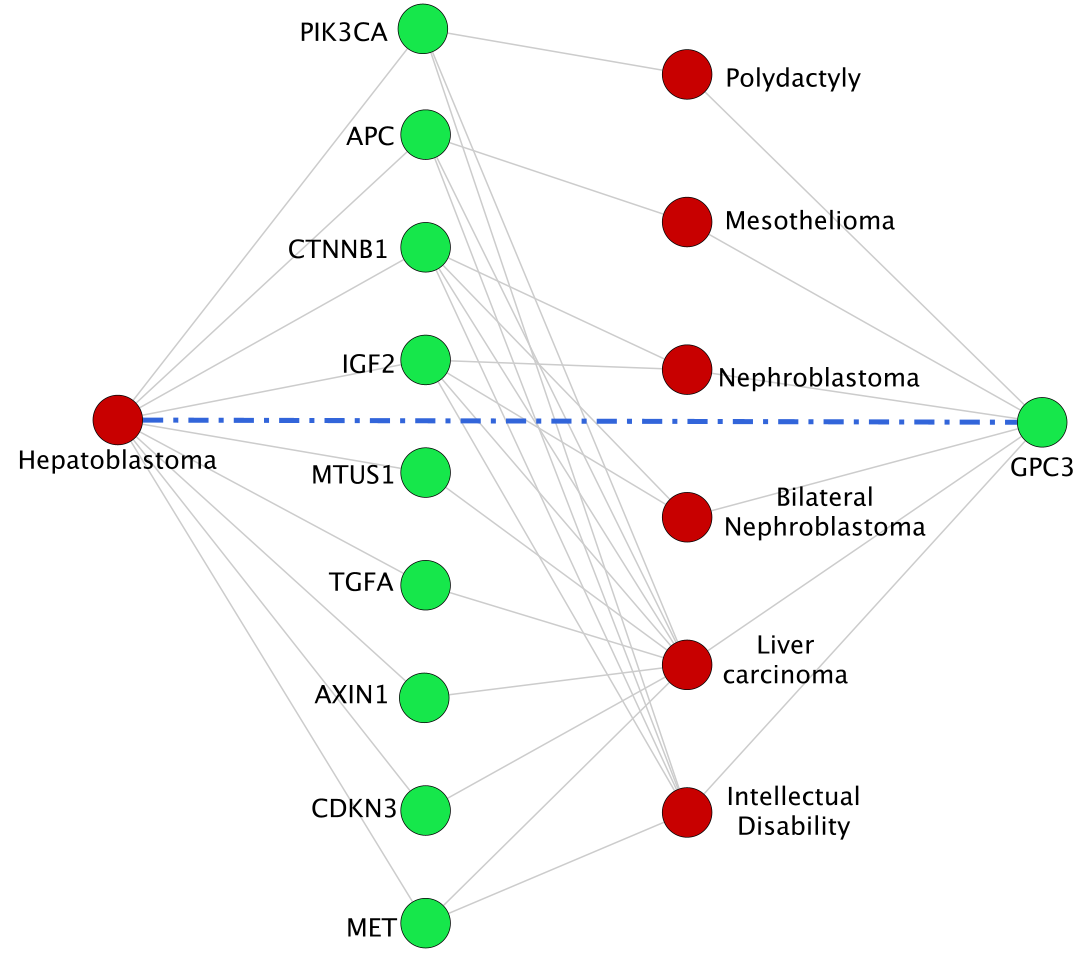}%
\label{fig_gpc3_hepatoblastoma}}
\hfil
\label{fig_network_gdi}
\caption{Subnetwork with predicted interactions (marked by bold dashed lines) between (a) ABO and Pancreatic carcinoma (b) GPC3 and Hepatoblastoma and all shortest paths between these pairs. The known interactions are presented as gray lines. Diseases are presented as dark circles and genes are presented as white circles.}
\end{figure*}

We select two predicted GDIs with a large number of supporting evidence and investigate the reason for the improvement in predicted confidence with HOGCN. Specifically, we choose gene-disease pairs (a) ABO and Pancreatic carcinoma (26 pieces of evidence) and (b) GPC3 and Hepatoblastoma ((17 pieces of evidence). To explain the prediction, the subnetworks containing all shortest paths between these pairs are selected. In particular, there are 49 shortest paths of length 3 between ABO and Pancreatic carcinoma including 6 diseases and 15 genes (Fig.~\ref{fig_abo_pancreas}). Similarly, there are 20 shortest paths of length 3 between GPC3 and Hepatoblastoma including 6 diseases and 9 genes (Fig.~\ref{fig_gpc3_hepatoblastoma}). Since these nodes are 3-hop away from each other and GCNs can only consider immediate neighbors, GCNs assign low confidence to these interactions. 

Examining the subnetwork in Fig.~\ref{fig_abo_pancreas}, we found that most of the diseases are related to a cancerous tumor in the pancreas and the prostate. Furthermore, pancreatic carcinoma is associated with other diseases such as Pancreatic neoplasm, malignant neoplasm of pancreas, and malignant neoplasm of prostate~\cite{pinero2020disgenet}. Since ABO is linked with diseases that are related to pancreatic carcinoma and other genes are related to these diseases as well, HOGCN captures such association (Fig~\ref{fig_abo_pancreas}) even though they are farther away in the network. Similarly, HOGCN predicts association for GPC3 and Hepatoblastoma (Fig.~\ref{fig_gpc3_hepatoblastoma}). 

Next, we perform a similar case study for DDIs and evaluate the predictions against DrugBank~\cite{wishart2006drugbank}. For this experiment, we make a prediction for every drug pair with GCN, SkipGNN and HOGCN and exclude the interactions that are already in the training set. Table~\ref{tab:drug_drug_positives} shows the top 5 interactions with an increase in interaction probabilities when highers orders of the neighborhood are considered. As discussed in Section~\ref{calibration}, HOGCN makes predictions with higher confidence compared to GCN and SkipGNN for the interactions that are likely to be a true positive. 

\begin{table}[!htb]
    \centering
    \caption{Novel prediction of DDIs with the literature evidence supporting the interaction. GCN, SkipGNN and HOGCN are denoted by $1, 2$ and $3$ respectively.}
    \label{tab:drug_drug_positives}
\begin{tabular}{p{1.7cm}p{1.7cm}p{0.5cm}p{0.5cm}p{0.5cm}c}
\hline
\multirow{2}{*}{Drug 1} & \multirow{2}{*}{Drug 2}  & \multicolumn{3}{c}{Probability} &  \\
\cline{3-5}
 & & $1$ & $2$ & $3$  & Evidence \\
\hline
Nelfinavir & Acenocoumarol & 0.192 & 0.318 & 0.417 & \cite{garcia1999sequential} \\
Praziquantel & Itraconazole &	0.609	& 0.721 & 	0.811 & \cite{perucca2006clinically} \\
Cisapride &	Droperidol &0.618	&0.725	&0.823 & \cite{michalets2000drug}  \\
Dapsone & Warfarin & 0.632	&0.720	&0.885	& \cite{truong2012probable} \\
Levofloxacin & Tobramycin &0.663	&0.760	&0.823	& \cite{ozbek2009post}  \\
\hline 
\end{tabular}
\end{table}

Moreover, we validate the false positive DDI predictions of GCNs and investigate the subnetwork for these drugs in DDI networks to reason the predictions. Table~\ref{tab:drug_drug_negatives} shows the top 5 interactions with a significant decrease in predicted confidence compared to GCN-based models. Since these DDIs are false positives~\cite{wishart2006drugbank}, GCN-based models make overconfident predictions for such DDIs. In contrast, HOGCN significantly reduces the predicted confidence for these DDIs to be true positive, indicating that the higher-order neighborhood allows HOGCN to identify false positive predictions. In particular, HOGCN can identity false positive DDI between Belimumab and Estazolam even though they are 3-hop away from each other.

\begin{table}[!htb]
    \centering
    \caption{Predicted probability for negative DDIs. GCN, SkipGNN and HOGCN are denoted by $1, 2$ and $3$ respectively. }
    \label{tab:drug_drug_negatives}
\begin{tabular}{p{2.5cm}p{2.5cm}p{0.5cm}p{0.5cm}p{0.5cm}}
\hline
\multirow{2}{*}{Drug 1} & \multirow{2}{*}{Drug 2} & \multicolumn{3}{c}{Probability with $k$} \\
\cline{3-5}
& & $1$ & $2$ & $3$ \\
\hline
Tranylcypromine &	Melphalan&	0.925&	0.478&	0.065\\
Belimumab&	Estazolam&	0.912&	0.477&	0.178\\
Methotrimeprazine&	Cloxacillin&	0.907&	0.406&	0.065\\
Hydrocodone	&Melphalan&	0.905&	0.193&	0.012\\
Ibrutinib&	Mecamylamine&	0.899&	0.398&	0.353\\
\hline
\end{tabular}
\end{table}

\begin{figure}[htb]
\centering
\subfloat[]{\includegraphics[width=0.48\linewidth]{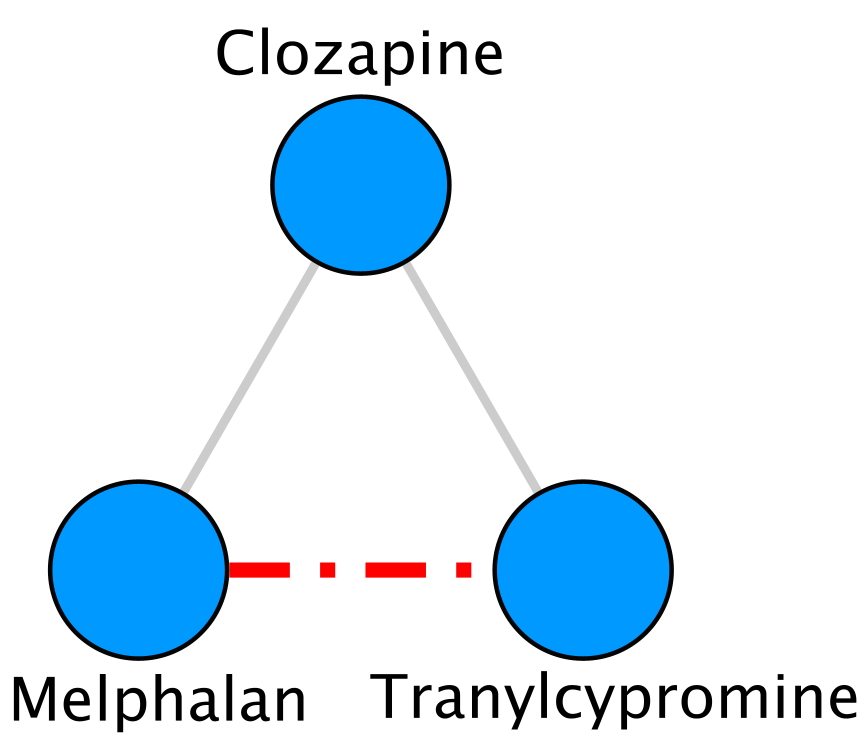}%
\label{fig_ddi_1}}
\hfil
\subfloat[]{\includegraphics[width=0.48\linewidth]{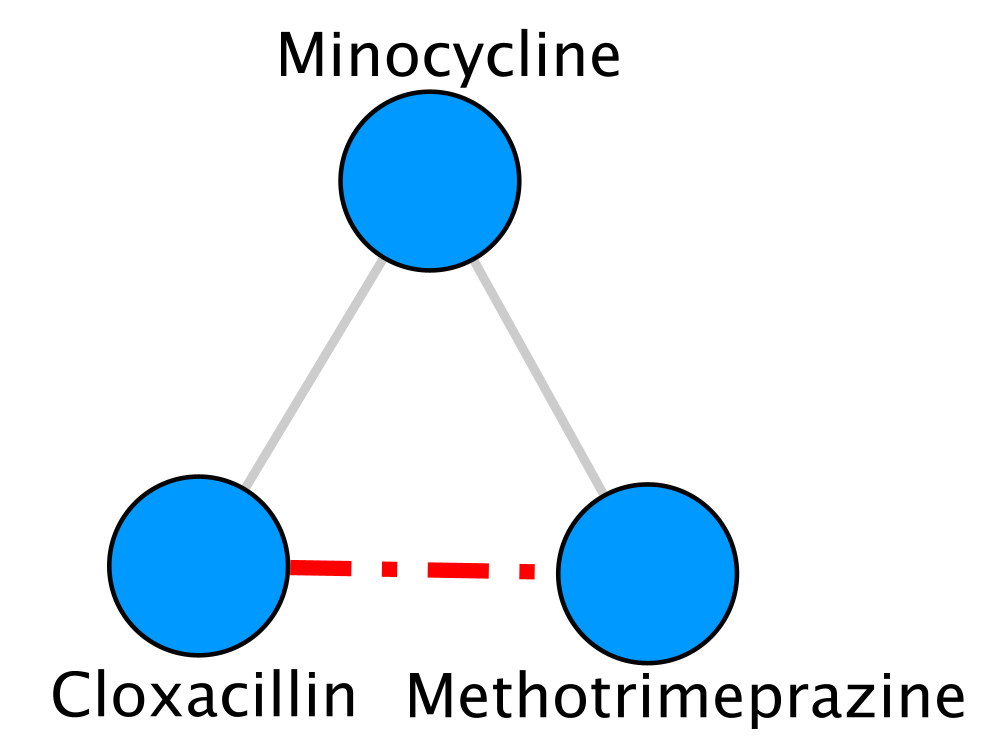}%
\label{fig_ddi_2}}
\hfil
\subfloat[]{\includegraphics[width=0.48\linewidth]{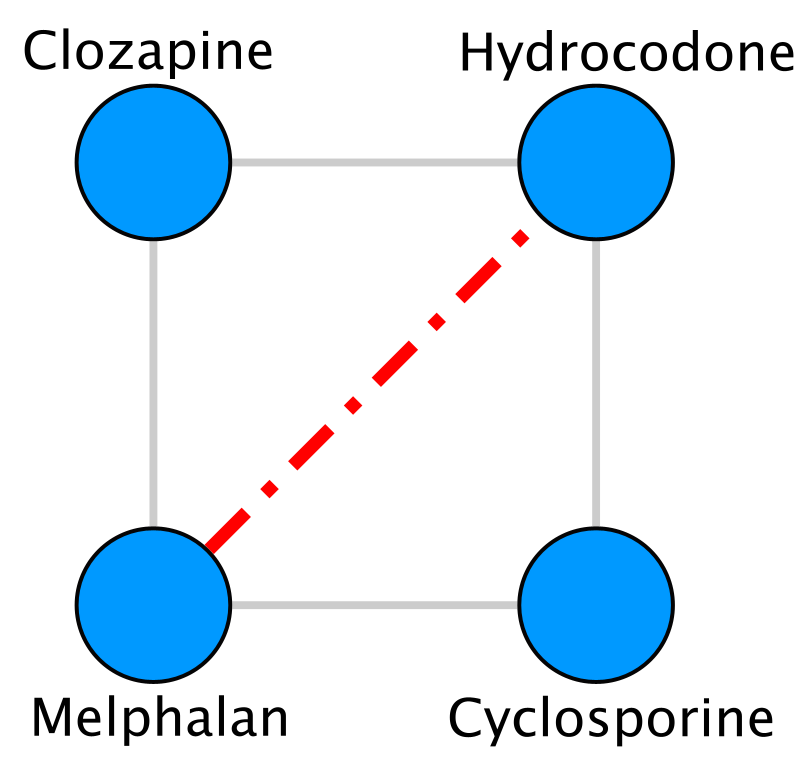}%
\label{fig_ddi_3}}
\hfil
\subfloat[]{\includegraphics[width=0.48\linewidth]{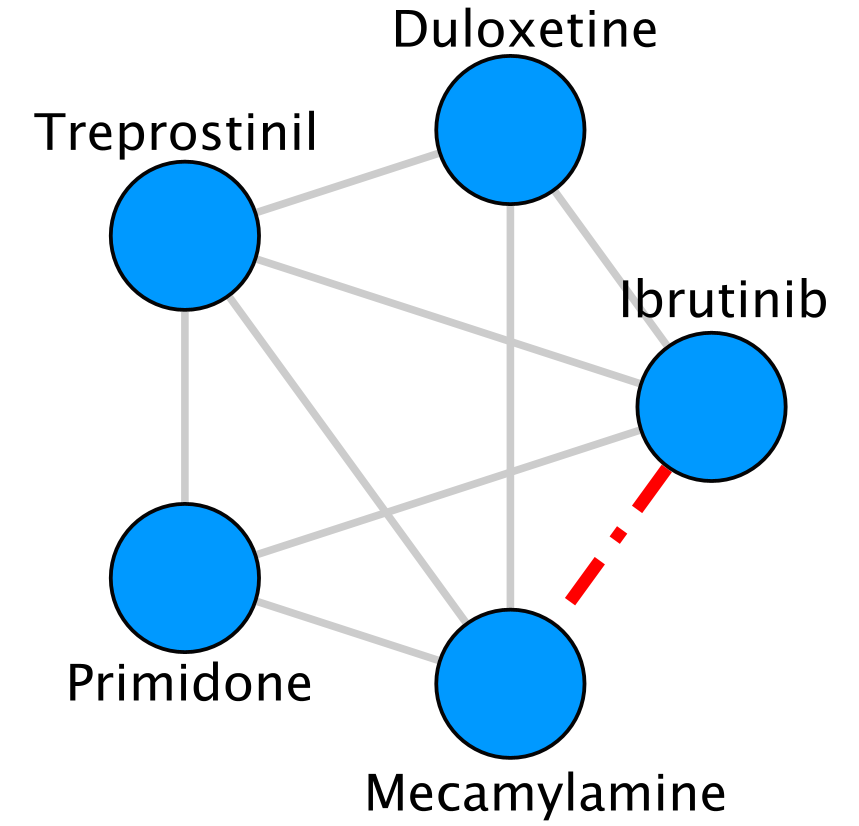}%
\label{fig_ddi_4}}
\hfil
\caption{Subnetwork containing false positive predictions (marked by dark dashed lines) and all shortest paths between (a) Tranylcypromine and Melphalan (b) Methotrimeprazine and Cloxacillin (c) Hydrocodone and Melphalan and (d) Ibrutinib and Mecamylamine. Other known interactions are presented as gray gray lines. Dark circles denotes drugs.}
\label{fig_network_ddi}
\end{figure}

We select subnetwork involving the drugs to investigate the reason for such predictions. Fig.~\ref{fig_network_ddi} shows the subnetwork with all shortest paths between the drugs in Table~\ref{tab:drug_drug_negatives}. Examining the figure, we observe that the drugs in these false positive DDIs have common immediate neighbors for all cases. GCN makes wrong predictions for these DDIs with high confidence. However, SkipGNN becomes less confident about the interaction being true positive by considering the skip similarity. HOGCN further reduces the predicted confidence for Tranylcypromine and Melphalan to 0.065, indicating that there is no association between these drugs. 

These case studies show that HOGCN with higher-order neighborhood mixing not only provide information for the identification of novel interactions but also help HOGCN to reduce false positive predictions.

\section{Conclusion}
We present a novel deep graph convolutional network (HOGCN) for biomedical interaction prediction. Our proposed model adopts a higher-order graph convolutional layer to learn to mix the feature representation of neighbors at various scales. Experimental results on four interaction datasets demonstrate the superior and robust performance of the proposed model. Furthermore, we show that HOGCN makes accurate and calibrated predictions by considering higher-order neighborhood information.

There are several directions for future study. Our approach only considers the known interactions to flag potential interactions. There are other sources of biomedical information such as various physicochemical and biological properties of biomedical entities that can provide additional information about the interaction and we plan to investigate the integration of such features into the model. As HOGCN aggregates the neighborhood information at various distances and can flag novel interactions, it would be interesting to provide interpretable explanations for the predictions in the form of a small subgraph of the input interaction network $\mathcal{G}$ that are most influential for the predictions~\cite{ying2019gnnexplainer}.

%

\ifCLASSOPTIONcompsoc
  \section*{Acknowledgments}
\else
  \section*{Acknowledgment}
\fi
This work was supported by the NSF [NSF-1062422 to A.H.], [NSF-1850492 to R.L.] and the NIH [GM116102 to F.C.]


\ifCLASSOPTIONcaptionsoff
  \newpage
\fi



\bibliographystyle{IEEEtran}
\bibliography{IEEEtran}




\end{document}